\documentclass[letterpaper,journal]{IEEEtran}
\usepackage{amsmath,amsfonts}
\usepackage{algorithmic}
\usepackage{algorithm}
\usepackage{array}
\usepackage[caption=false,font=normalsize,labelfont=sf,textfont=sf]{subfig}
\usepackage{textcomp}
\usepackage{stfloats}
\usepackage{url}
\usepackage{adjustbox}
\usepackage{verbatim}
\usepackage{multirow}
\usepackage{hyperref}
\usepackage[table]{xcolor}
\usepackage{graphicx}
\usepackage{amssymb}
\usepackage{cite}
\usepackage{amsthm}

\usepackage{bm}
\usepackage{booktabs}
\usepackage{makecell}
\definecolor{PairFirst}{RGB}{198,239,206}   
\definecolor{PairSecond}{RGB}{189,215,238}  
\definecolor{SceneFirst}{RGB}{255,199,206}  
\definecolor{SceneSecond}{RGB}{255,235,156} 

\newcommand{\pairfirst}[1]{\cellcolor{PairFirst}#1}
\newcommand{\pairsecond}[1]{\cellcolor{PairSecond}#1}
\newcommand{\scenefirst}[1]{\cellcolor{SceneFirst}#1}
\newcommand{\scenesecond}[1]{\cellcolor{SceneSecond}#1}

\makeatletter
\renewcommand{\@IEEEtablecaptionsepspace}{}
\makeatother

\begin{document}

\title{From Pair to Scene: Free-View Infrared–Visible Fusion from Misaligned RGB–IR Observations}

\author{Shengjie Hu, ~Xiaogang Zhang\textsuperscript{*} \IEEEauthorblockA{\emph{Member, IEEE}}, ~Hua Chen\textsuperscript{*}, ~Zhengzhao Pan, ~Xiaoyu Zhu, ~and Jiuye Shi 
\thanks{This work was funded by the National Natural Science Foundation of China with the project approval numbers: U23A20385, 62273139 and 62377010.
\textsuperscript{*}Corresponding author.

Shengjie Hu and Xiaoyu Zhu are with the school of Artificial and Intelligence and Robotics, Hunan University, Changsha, Hunan 410000, China (e-mail: hnuhsj1230@hnu.edu.cn; xzhu5@hnu.edu.cn).

Xiaogang Zhang, Zhengzhao Pan and Jiuye Shi are with the College of Electrical and Information Engineering, Hunan University, Changsha, Hunan 410000,China (e-mail: zhangxg@hnu.edu.cn; zzpan@hnu.edu.cn; shi948880415@gmail.com).

Hua Chen is with the College of Computer Science and Electronic Engineering, Hunan University, Changsha, Hunan 410000, China (e-mail:chua@hnu.edu.cn).}
}

\maketitle

\begin{abstract}
Pairwise infrared--visible image fusion (IVF) combines visible and thermal information at captured viewpoints but does not directly support fused rendering along user-defined camera trajectories. We study free-view IVF as a scene-level task that synthesizes fused images at arbitrary user-defined viewpoints from approximately paired but residually misaligned multi-view RGB--IR observations of a static scene. Joint optimization faces two coupled challenges: undesirable co-adaptation between registration and scene representation, and the tension between accommodating modality-dependent spatial responses and maintaining the cross-modal correspondence needed for fusion. We propose Beyond3D, a coarse-to-fine scene-level fusion framework that jointly optimizes registration, a 3D Gaussian scene representation, and view-conditioned fusion weights to learn a fusion field. The coarse stage establishes shared geometry through alternating registration and representation updates, with fixed-reference consensus constraining their co-adaptation. The fine stage introduces paired RGB and IR children with bounded spatial variation around canonical parent Gaussians to accommodate modality-dependent spatial responses. Gaussian- and ray-level correspondence constraints regularize the parent-defined cross-modal association needed for fusion. View-conditioned pre-projection fusion combines the paired appearance attributes before a single rendering pass over the canonical parent geometry. Reliability-constrained registration and RGB-protected geometry updates further stabilize refinement. Experiments demonstrate competitive fusion quality at held-out test viewpoints and coherent fused rendering along the evaluated camera trajectories. The code and data will be publicly available at \url{https://github.com/PopC4N/FreeIVF-3D}.
\end{abstract}


\section{Introduction}
\begin{figure}[t]
\centering
\includegraphics[width=1.0\columnwidth]{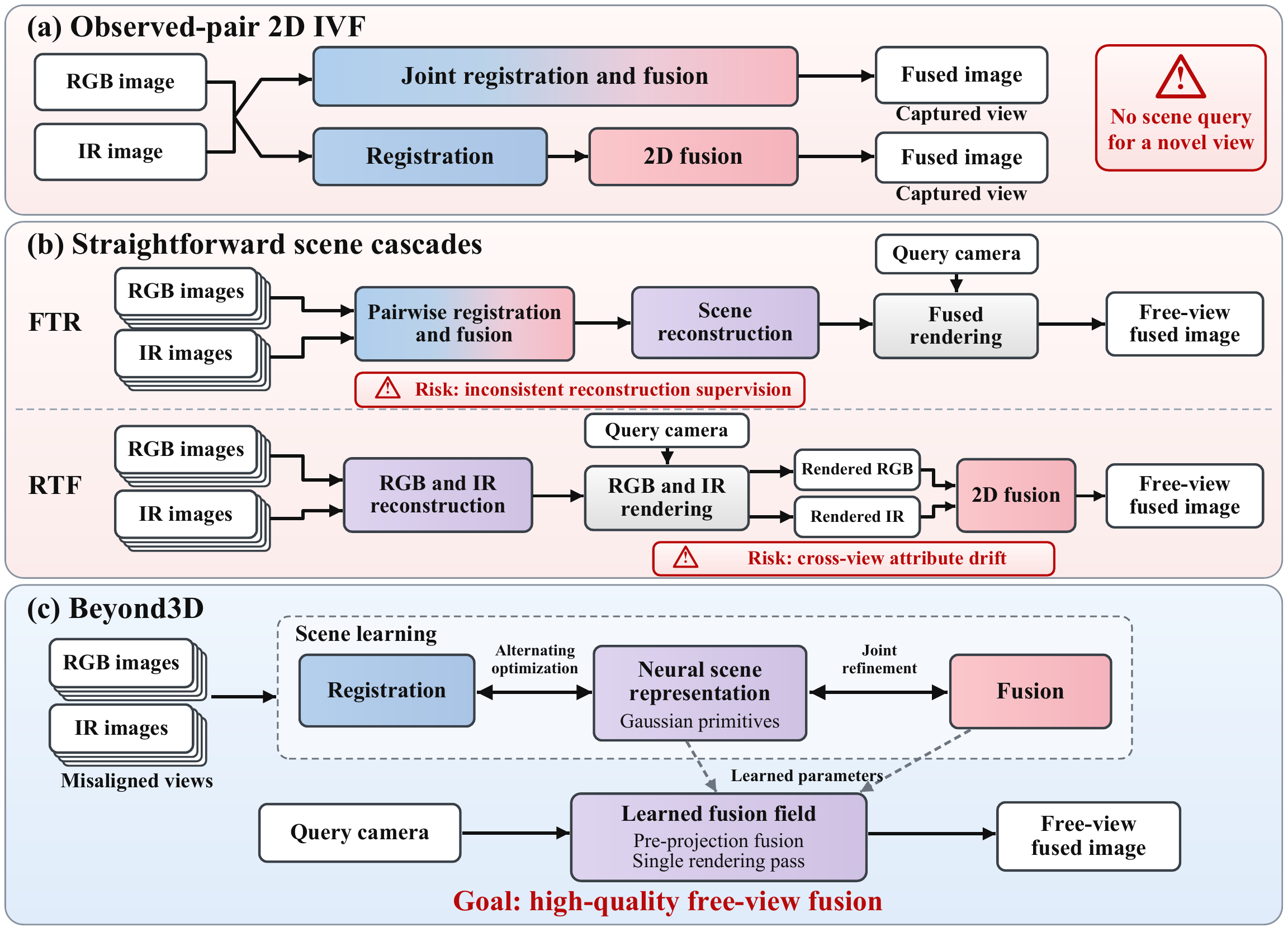}
\caption{Conceptual comparison of infrared-visible fusion formulations.
(a) Pairwise IVF methods fuse captured views without a scene-query interface.
(b) Straightforward scene cascades enable free-view fusion but risk inconsistent reconstruction supervision or cross-view attribute drift.
(c) Beyond3D couples registration and fusion through a Gaussian scene representation, enabling camera queries via pre-projection fusion. Bidirectional arrows indicate feedback during learning.}
\label{fig:1}
\end{figure}

\IEEEPARstart{I}{nfrared--visible} image fusion (IVF) combines complementary visible details and thermal infrared cues in a single image~\cite{ma2019infrared}. Most IVF methods operate on an observed RGB--IR pair to support visual inspection or downstream perception~\cite{li2018densefuse,zhao2023cddfuse,tang2022image}, as illustrated in Fig.~\ref{fig:1}(a). However, in aerial inspection, robotic scanning, and intelligent monitoring, users may need to examine a thermal anomaly from an uncaptured viewpoint or inspect its relationship to visible structures along a continuous virtual camera trajectory. Moreover, in practical dual-spectrum systems, RGB--IR observations often exhibit residual misalignment at the pixel level due to calibration and mounting errors, small-baseline viewpoint discrepancies, and field-of-view or acquisition variations~\cite{xu2023murf,wang2024improving}. We therefore study an important yet underexplored task, termed free-view IVF: given approximately paired but residually misaligned multi-view RGB--IR observations of a static scene, together with camera poses estimated from the RGB images using structure from motion (SfM)~\cite{schonberger2016structure}, the goal is to synthesize coherent fused images at arbitrary user-defined viewpoints.

Existing pairwise IVF methods cannot directly fulfill this task because they require source images at the requested viewpoint. Neural scene representations~\cite{mildenhall2021nerf,kerbl20233d} enable scene reconstruction from multi-view observations and subsequent rendering at queried viewpoints. A straightforward approach is to combine neural scene representations with pairwise IVF methods in a pair-to-scene cascade, as illustrated in Fig.~\ref{fig:1}(b). Fusion-then-reconstruction (FTR) registers and fuses each training RGB--IR pair, then reconstructs a scene from the resulting fused images. Residual misalignment and differences in fusion across pairs can make these images inconsistent observations of the same scene, causing reconstruction to fit conflicting structural and appearance information. Reconstruction-then-fusion (RTF) instead registers the training observations, reconstructs both modalities, renders them at a query viewpoint, and applies pairwise fusion to the rendered pair. Although reconstruction organizes the observations in scene space, the subsequent fusion stage does not explicitly constrain how the same scene region is fused across views. Its fused appearance may therefore change inconsistently as the queried viewpoint varies.

These limitations arise because both cascades optimize fusion and reconstruction separately rather than in a unified manner. Direct joint optimization faces two coupled challenges. First, joint optimization can induce undesirable co-adaptation between registration and scene representation. The scene representation may absorb registration errors by adapting its spatial structure to fit misaligned observations. The resulting representation may in turn favor the same biased registration. The reconstruction loss can therefore decrease without correcting the spatial correspondence. Second, the scene representation should accommodate modality-dependent spatial responses while maintaining the cross-modal correspondence needed for fusion. Even after accurate alignment, differences in visibility and spatial support may not be fully captured by shared geometry with separate appearance parameters~\cite{lin2024thermalnerf}. Allowing modality-specific geometry to fit the observations independently does not ensure that the appearance attributes to be fused describe the same local scene region, even when both modalities are reconstructed accurately.

To address these challenges, we propose Beyond3D, a coarse-to-fine scene-level fusion framework that learns a fusion field from residually misaligned multi-view RGB--IR observations and camera poses estimated using SfM, as depicted in Fig.~\ref{fig:1}(c). The fusion field is parameterized by a 3D Gaussian representation and view-conditioned fusion weights. Following per-view registration initialization, the coarse stage limits the mutual reinforcement of registration and representation errors through alternating updates and fixed-reference consensus (FRC) under shared Gaussian geometry. To accommodate modality-dependent spatial responses while retaining a common spatial basis for fusion, the fine stage introduces paired RGB and IR children with bounded spatial variation around canonical parent Gaussians. Gaussian- and ray-level correspondence constraints regularize the parent-defined cross-modal association before rendering and through image formation, encouraging the paired children to explain the same local scene region during modality-specific refinement. View-conditioned fusion weights combine the paired appearance attributes before a single rendering pass over the canonical parent geometry, enabling fused rendering from a query viewpoint after optimization. Reliability-constrained registration limits unsupported alignment updates, while RGB-protected geometry updates reduce conflicts between RGB and IR reconstruction gradients on the shared parent geometry. Figure~\ref{fig:2} summarizes the training and inference pipeline.

The main contributions of this work are summarized as follows:

(1) We study the important yet underexplored task of free-view IVF and formulate it at the scene level, extending fusion beyond captured image pairs to arbitrary user-defined viewpoints. The formulation uses approximately paired but residually misaligned multi-view RGB--IR observations of a static scene to synthesize coherent fused images, without requiring precisely aligned input pairs or an observed RGB--IR pair at the query viewpoint.

(2) We develop a coarse-to-fine scene-level fusion framework that jointly refines residual cross-modal registration and a 3D Gaussian scene representation. Alternating optimization and fixed-reference consensus suppress co-adaptation between registration and representation.

(3) We introduce paired RGB and IR children with bounded spatial variation around canonical parent Gaussians to accommodate modality-dependent spatial responses, while Gaussian- and ray-level correspondence constraints regularize the parent-defined cross-modal association needed for fusion. View-conditioned pre-projection fusion combines the paired appearance attributes before a single rendering pass over the canonical parent geometry.

Extensive experiments demonstrate competitive fusion quality at held-out test viewpoints relative to representative pairwise IVF methods and scene-level cascade baselines, together with coherent fused rendering along the evaluated camera trajectories. Comprehensive ablation studies further validate the effectiveness of the proposed components.

\section{Related Work}
\label{sec:related-work}

\subsection{Infrared and Visible Image Fusion}

Infrared and visible image fusion seeks to retain thermal cues and visible structure within a common image~\cite{ma2019infrared}. For registered inputs, reconstruction features support explicit aggregation, while feature decomposition separates correlated base features from complementary details~\cite{li2018densefuse,zhao2023cddfuse}. Without a fused reference, source similarity constraints guide information retention~\cite{xu2020u2fusion}. Adversarial learning combines image statistics with content constraints~\cite{ma2019fusiongan,liu2022target}, while masked conditional diffusion learns a generative prior from reconstruction tasks~\cite{tang2025mask}. 
When the inputs are misaligned, correspondence becomes part of the fusion task~\cite{xu2023murf,wang2024improving,xiao2026hybrid}.
For visible and infrared images, fused-gradient consistency and sparsity have been used to guide fine registration~\cite{xu2022rfnet}. Modality translation reduces the difficulty of cross-modal matching~\cite{wang2022unsupervised,wang2024improving}, while shared structural representations support registration and fusion in a common feature space~\cite{xiao2026hybrid}. Correction can also follow an initial fusion step, with localized warping rectifying misaligned content in the fused result~\cite{bian2026fusionregister}. 

Our work extends this coupling to a persistent, camera-queryable scene. It jointly refines residual registration, multimodal Gaussian representation, and a view-conditioned pre-projection fusion field. Correspondence therefore associates visible and thermal attributes within the scene, enabling their fusion across camera queries without requiring a source image pair at the target viewpoint. This changes both the domain in which fusion is defined and the input required to obtain a fused image.

\subsection{Neural Scene Representation}

Scene representations support camera queries by explaining multi-view observations within a common spatial model. Neural radiance fields (NeRF) encode continuous density and directional radiance~\cite{mildenhall2021nerf}. Tensor factorization, sparse voxel fields, and multiresolution hash features improve the efficiency of scene representation and optimization~\cite{chen2022tensorf,fridovich2022plenoxels,muller2022instant}, while 3D Gaussian Splatting (3DGS) uses explicit primitives and differentiable rasterization for efficient rendering~\cite{kerbl20233d}. These formulations provide a computational basis for free-view IVF, whose multimodal setting additionally requires a choice of how RGB and IR should share spatial structure.

Multimodal representations balance common scene structure against differences in spectral response. The OMMG variant of ThermalGaussian uses shared geometry and opacity with separate appearance coefficients~\cite{lu2026thermalgaussian++}. More flexible spatial support can be modeled through separate density fields with regularized differences, as in ThermalNeRF~\cite{lin2024thermalnerf}, or through shared and modality-specific Gaussian primitives, as in MMOne~\cite{gu2025mmone}. Our method likewise separates shared structure from modality differences, but retains a canonical parent for each paired RGB and IR child. Bounded geometric residuals accommodate local modality differences, while parent-defined correspondence supplies the paired attributes and common spatial support used for fused rendering.

Alignment introduces a further distinction between establishing a common coordinate frame and refining residual image correspondence. Cross-spectral reconstruction can use camera calibration or pose alignment~\cite{wang2026binar,cordonnier2026unpaired}, and can refine cross-modal alignment together with the scene~\cite{poggi2022cross,guo2025cross,lu2026thermalgaussian++}. Joint refinement of alignment and scene representation is therefore already established. Our method addresses residual image misregistration in approximately paired RGB-IR observations through per-view homography refinement. Because registration and scene geometry can both explain image discrepancies, it alternates their updates and constrains modality-specific geometric freedom. The resulting representation also defines a view-conditioned fusion field: paired appearance attributes are mixed before a single projection and rendering pass on canonical parents. The distinction thus lies in coupling residual registration and constrained modality geometry with a common support for free-view fused rendering.

\begin{figure*}[t]
\centering
\includegraphics[width=1.0\textwidth]{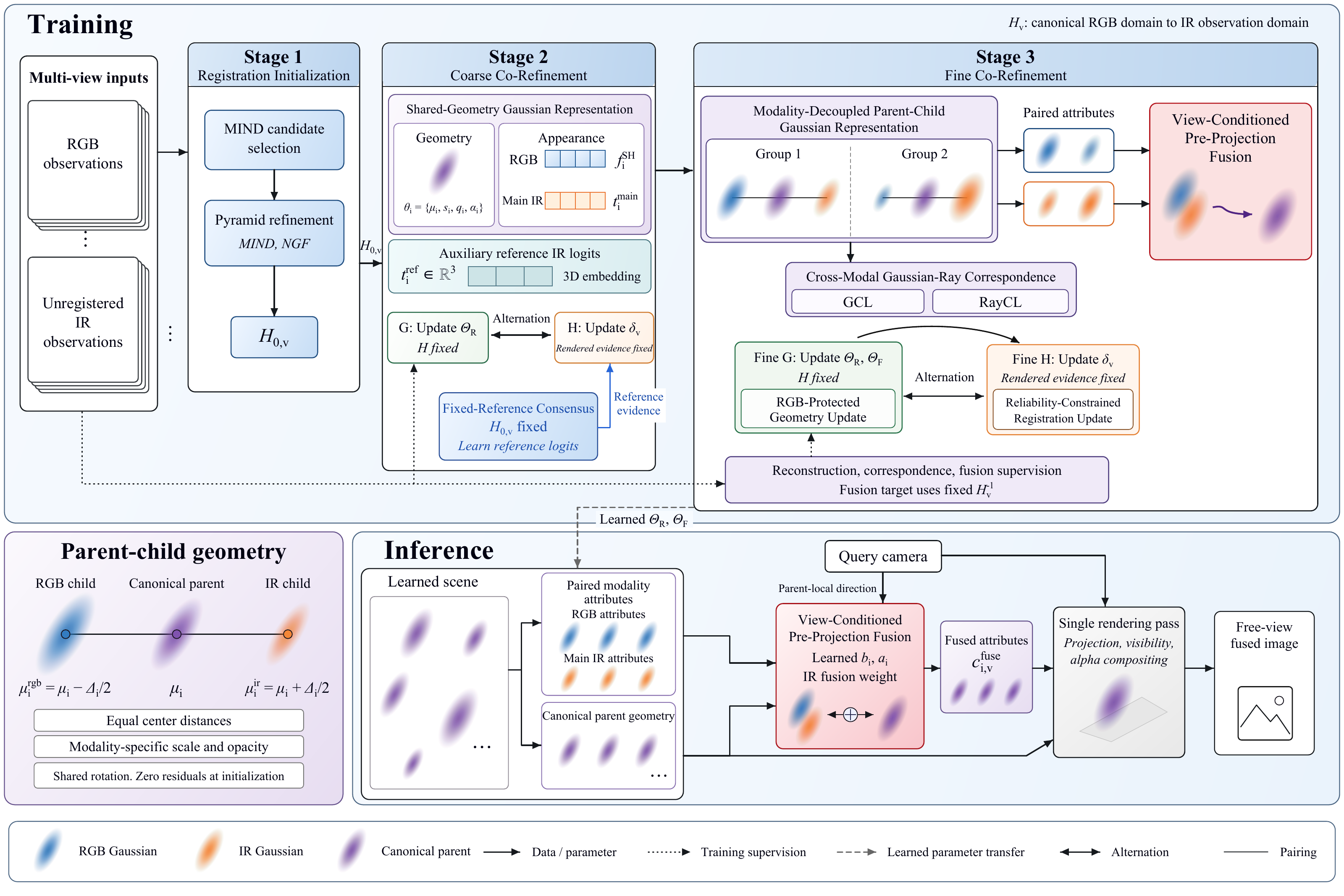}
\caption{Overview of Beyond3D. Training comprises per-view registration initialization, coarse co-refinement with shared Gaussian geometry, and fine co-refinement with paired modality-specific children and view-conditioned pre-projection fusion. Registration and scene optimization alternate, while Gaussian- and ray-level losses regularize parent-defined correspondence. At inference, paired modality attributes are fused before a single rendering pass over canonical parent geometry, using only the query camera and learned scene parameters.}
\label{fig:2}
\end{figure*}

\section{Methodology}
\label{sec:method}

\subsection{Overall Framework}
We take as input multi-view RGB-IR observations $\{I_v^{\mathrm{rgb}},I_{v,u}^{\mathrm{ir}}\}_{v=1}^{V}$ of a static scene and the associated camera poses estimated from the RGB images using structure from motion \cite{schonberger2016structure}. Each RGB-IR pair has approximate correspondence with residual cross-modal misalignment. Our goal is to learn a camera-queryable scene representation for free-view fused rendering. The proposed framework jointly optimizes the per-view homographies $\{H_v\}$ for residual cross-modal image registration, the scene representation parameters $\Theta_R$, and the fusion parameters $\Theta_F$.

As illustrated in Fig. \ref{fig:2}, the framework consists of three components: per-view cross-modal registration initialization, coarse registration--representation co-refinement, and fine registration--representation--fusion co-refinement.
The initialization provides reliable view-specific transformations.
The coarse optimization establishes a canonical Gaussian representation under shared geometry.
The fine optimization further introduces modality-decoupled representation and jointly refines registration and fusion.
After optimization, the learned representation supports free-view fused rendering at arbitrary viewpoints.

\subsection{Per-View Cross-Modal Registration Initialization}

Before optimizing the Gaussian representation, we independently initialize a cross-modal homography for each training view in the observation domain.
For any homography $H$, its warp-induced valid support and overlap ratio are defined as
\begin{equation}
\begin{aligned}
M_v(H)
&=
\mathbf 1
\left[
\operatorname{warp}(\mathbf 1,H)>0.999
\right],
\\
o_v(H)
&=
\operatorname{mean}_{p}M_v(H,p).
\end{aligned}
\label{eq:init_valid_support}
\end{equation}

For each view $v$, we sample $K$ candidate homographies $\{H_{v,k}^{\mathrm{cand}}\}_{k=1}^{K}$.
Let $D_v^{\mathrm{rgb}}$ and $D_{v,u}^{\mathrm{ir}}$ denote the MIND descriptors~\cite{heinrich2012mind} of the RGB and unregistered IR observations, respectively. 
The valid support and structural matching score of candidate $k$ are
\begin{equation}
\begin{aligned}
M_{v,k}
&=
M_v
\left(
H_{v,k}^{\mathrm{cand}}
\right),
\\
s_{v,k}^{\mathrm{mind}}
&=
\operatorname{mean}_{M_{v,k},c}
\left|
\operatorname{warp}
\left(
D_v^{\mathrm{rgb}},
H_{v,k}^{\mathrm{cand}}
\right)
-
D_{v,u}^{\mathrm{ir}}
\right|,
\end{aligned}
\label{eq:init_candidate_score}
\end{equation}
where the masked mean is taken over valid pixels and descriptor channels.
The candidate with the lowest score is selected as the base estimate:
\begin{equation}
k_v^{\star}
=
\arg\min_{1\leq k\leq K}
s_{v,k}^{\mathrm{mind}},
\qquad
H_v^{\mathrm{base}}
=
H_{v,k_v^{\star}}^{\mathrm{cand}}.
\label{eq:init_candidate_selection}
\end{equation}

Starting from $H_v^{\mathrm{base}}$, we optimize an initialization-specific eight-dimensional residual $\boldsymbol\eta_v\in\mathbb R^8$ over a coarse-to-fine image pyramid. The corresponding homography is parameterized as
\begin{equation}
H_v^{\mathrm{init}}
\left(
\boldsymbol\eta_v
\right)
=
\exp
\left(
A(\boldsymbol\eta_v)
\right)
H_v^{\mathrm{base}},
\qquad
A(\boldsymbol\eta_v)
\in
\mathfrak{sl}(3).
\label{eq:init_residual_parameterization}
\end{equation}
The initialization objective is
\begin{equation}
\begin{aligned}
\mathcal L_{\mathrm{init}}^{(v)}
\left(
\boldsymbol\eta_v
\right)
&=
\mathcal L_{\mathrm{mind}}^{(v)}
\left(
\boldsymbol\eta_v
\right)
+
\lambda_{\mathrm{ngf}}
\mathcal L_{\mathrm{ngf}}^{(v)}
\left(
\boldsymbol\eta_v
\right)
+
\lambda_{\eta}
\left\|
\boldsymbol\eta_v
\right\|_2^2
\\
&\quad+
\left[
\max
\left(
0,
\tau_{\mathrm{init}}
-
o_v
\left(
H_v^{\mathrm{init}}
\left(
\boldsymbol\eta_v
\right)
\right)
\right)
\right]^2.
\end{aligned}
\label{eq:registration_initialization}
\end{equation}
The MIND and NGF terms are evaluated over the valid support $M_v(H_v^{\mathrm{init}}(\boldsymbol\eta_v))$.
MIND provides the primary cross-modal structural constraint, while the sign-insensitive NGF~\cite{haber2006intensity} term complements it by emphasizing boundary alignment under possible gradient-polarity reversal. The residual regularizer and overlap penalty suppress unsupported projective deformation and degenerate low-overlap solutions, respectively.

The optimized residual, initialized homography, and final initialization loss are
\begin{equation}
\begin{aligned}
\boldsymbol\eta_v^{\star}
&=
\arg\min_{\boldsymbol\eta_v}
\mathcal L_{\mathrm{init}}^{(v)}
\left(
\boldsymbol\eta_v
\right),
\\
H_{0,v}
&=
H_v^{\mathrm{init}}
\left(
\boldsymbol\eta_v^{\star}
\right),
\\
\ell_v^{\mathrm{init}}
&=
\mathcal L_{\mathrm{init}}^{(v)}
\left(
\boldsymbol\eta_v^{\star}
\right).
\end{aligned}
\label{eq:init_outputs}
\end{equation}
The resulting $H_{0,v}$ initializes coarse co-refinement, while $\ell_v^{\mathrm{init}}$ provides the scene-relative quality statistic used for subsequent view-wise IR reliability weighting.

\subsection{Coarse Registration--Representation Co-Refinement}
Starting from the initialized homographies $\{H_{0,v}\}_{v=1}^{V}$, coarse co-refinement optimizes view-dependent homographies and a shared-geometry Gaussian scene at half resolution.
RGB and IR share Gaussian geometry and topology while retaining independent appearance parameters, excluding modality-specific geometric degrees of freedom that could absorb residual misregistration.
Alternating optimization removes within-block update coupling by holding the homographies fixed during scene updates and the representation fixed during homography updates.
The main representation can nevertheless adapt to biased homographies across successive blocks.
Fixed-reference consensus (FRC) further constrains this co-adaptation by evaluating homography updates against an auxiliary IR appearance model learned under the fixed initialized homographies.

\subsubsection{Shared-Geometry Gaussian Representation}

We represent RGB and IR using a common Gaussian set $\mathcal G=\{\mathcal G_i\}_{i=1}^{N}$ with shared geometry and modality-specific appearance. The $i$-th Gaussian is parameterized as
\begin{equation}
\mathcal G_i
=
\left\{
\underbrace{
\boldsymbol\mu_i,
\mathbf s_i,
\mathbf q_i,
\alpha_i
}_{\boldsymbol\theta_i},
\mathbf f_i^{\mathrm{SH}},
\mathbf t_i^{\mathrm{main}}
\right\},
\label{eq:coarse_gaussian}
\end{equation}
where $\boldsymbol\theta_i$ comprises the center, scale, rotation, and opacity. 
Sharing $\boldsymbol\theta_i$ gives the two modalities a common spatial support, visibility ordering, and Gaussian topology. 
Their appearance is represented independently by the RGB spherical-harmonic coefficients $\mathbf f_i^{\mathrm{SH}}$ and the main IR logits $\mathbf t_i^{\mathrm{main}}\in\mathbb R^3$. 
The coarse representation parameters are
\begin{equation}
\Theta_R
=
\left\{
\boldsymbol\theta_i,
\mathbf f_i^{\mathrm{SH}},
\mathbf t_i^{\mathrm{main}}
\right\}_{i=1}^{N}.
\label{eq:coarse_representation_parameters}
\end{equation}

For view $v$, the RGB appearance is decoded as $\mathbf c_{i,v}^{\mathrm{rgb}} =\operatorname{SH}(\mathbf f_i^{\mathrm{SH}},\mathbf d_{i,v})$, whereas the main IR appearance is parameterized by the view-independent color $\mathbf c_i^{\mathrm{main}} =\sigma(\mathbf t_i^{\mathrm{main}})$. 
The latter avoids unnecessary directional degrees of freedom that could absorb residual geometric error.
Rasterizing the two appearance attributes over the shared geometry yields $R_v^{\mathrm{rgb}}$ and $R_v^{\mathrm{main}}$ through front-to-back alpha compositing.

\subsubsection{Alternating Homography--Representation Optimization}
\label{sec:H/G}
The current homography of view $v$ is parameterized as a residual update around its initialized estimate:
\begin{equation}
H_v
=
\exp
\left(
A(\boldsymbol\delta_v)
\right)
H_{0,v},
\qquad
\boldsymbol\delta_v\in\mathbb R^8,
\label{eq:coarse_homography}
\end{equation}
where $A(\boldsymbol\delta_v)\in\mathfrak{sl}(3)$. Simultaneously updating $H_v$ and $\Theta_R$ creates \emph{moving-target supervision}, because both the rendered content and the warp used to compare it with the unregistered IR observation can change within the same backward pass. We therefore alternate between two block-coordinate updates:
\begin{equation}
\begin{aligned}
\Theta_R^{+}
&\leftarrow
\operatorname{Opt}_{G}
\left(
\Theta_R;
\operatorname{sg}(\mathbf H)
\right),
&& \text{G-block},
\\
\boldsymbol\delta_v^{+}
&\leftarrow
\operatorname{Opt}_{H}
\left(
\boldsymbol\delta_v;
\operatorname{sg}(\Theta_R)
\right),
&& \text{H-block},
\end{aligned}
\label{eq:coarse_alternation}
\end{equation}
where $\mathbf H=\{H_v\}_{v=1}^{V}$ and $\operatorname{sg}(\cdot)$ denotes stop-gradient. The G-block updates the representation under fixed registration, whereas the H-block refines one homography against frozen rendered evidence. Reliability is handled asymmetrically: uncertain IR observations are softly downweighted in the G-block, while H-block proposals are guarded by normalized multi-cue evidence and fixed-reference consensus.

In the G-block, the main representation is optimized using
\begin{equation}
\mathcal L_{\mathrm{G,main}}^{(v)}
=
\mathcal L_{\mathrm{rgb}}^{(v)}
+
\omega_v^{\mathrm{ir}}
\mathcal L_{\mathrm{ir,main}}^{(v)}.
\label{eq:coarse_main_objective}
\end{equation}

The fixed view-wise reliability weights are computed from the final initialization losses. 
Let $m_{\ell}=\operatorname{median}_{v}(\ell_v^{\mathrm{init}})$ and $\operatorname{MAD}_{\ell}=\operatorname{median}_{v}|\ell_v^{\mathrm{init}}-m_{\ell}|$, and define
$s_{\ell}=\max(1.4826\,\operatorname{MAD}_{\ell},10^{-4})$.
We then set $r_v=\sigma((m_{\ell}-\ell_v^{\mathrm{init}})/s_{\ell})$ and $\omega_v^{\mathrm{ir}}=0.1+0.9r_v$.
The weights remain fixed throughout coarse co-refinement, attenuating IR supervision from less reliable views without discarding them.

RGB supervision combines photometric and structural similarity~\cite{wang2004image}:
\begin{equation}
\begin{aligned}
\mathcal L_{\mathrm{rgb}}^{(v)}
&=
\lambda_{\mathrm{rgb}}
\left\|
R_v^{\mathrm{rgb}}
-
I_v^{\mathrm{rgb}}
\right\|_1
\\
&\quad+
\lambda_{\mathrm{ssim}}
\left[
1-
\operatorname{SSIM}
\left(
R_v^{\mathrm{rgb}},
I_v^{\mathrm{rgb}}
\right)
\right],
\end{aligned}
\label{eq:coarse_rgb_loss}
\end{equation}
where $\lambda_{\mathrm{rgb}}=0.8$ and $\lambda_{\mathrm{ssim}}=0.2$.

The valid support $M_v(H)$ and overlap ratio $o_v(H)$ follow Eq.~\eqref{eq:init_valid_support}. They depend only on warp-induced source support and not on the observed IR intensity, so dark or zero-valued IR pixels remain valid wherever source support exists. For a prediction $P$ warped by $H$ and a target $T$, we define
\begin{equation}
\begin{aligned}
\mathcal L_{\mathrm{obs}}^{(v)}
\left(
P,T;H
\right)
&=
\operatorname{mean}_{M_v(H),c}
\sqrt{
(P_c-T_c)^2+\epsilon_c
}
\\
&\quad+
\lambda_{\nabla}
\operatorname{mean}_{M_v(H)}
\sqrt{
\left\|
\nabla_{\mathrm S}P
-
\nabla_{\mathrm S}T
\right\|_2^2
+
\epsilon_c
}
\\
&\quad+
\left[
\max
\left(
0,
\tau_o-o_v(H)
\right)
\right]^2,
\end{aligned}
\label{eq:coarse_observation_loss}
\end{equation}
where $\epsilon_c=10^{-6}$, $\lambda_{\nabla}=0.2$, and $\tau_o=0.45$.
The first mean is taken over valid pixels and channels, while the second is taken over valid pixels. The Sobel gradient is computed on luminance $Y(I)=0.299I_R+0.587I_G+0.114I_B$ as $\nabla_{\mathrm S}I=(K_x\ast Y(I),K_y\ast Y(I))$.
The overlap term suppresses degenerate low-overlap transformations; it is constant with respect to $\Theta_R$ in the G-block because $H_v$ is detached.

The main IR reconstruction loss is therefore
\begin{equation}
\mathcal L_{\mathrm{ir,main}}^{(v)}
=
\mathcal L_{\mathrm{obs}}^{(v)}
\left(
\operatorname{warp}
\left(
R_v^{\mathrm{main}},H_v
\right),
I_{v,u}^{\mathrm{ir}};
H_v
\right).
\label{eq:coarse_main_ir_loss}
\end{equation}

In the H-block, all rendered evidence is detached and only $\boldsymbol\delta_v$ is optimized. The main IR and MIND registration cues are defined as functions of a candidate homography $H$:
\begin{equation}
\begin{aligned}
\ell_v^{\mathrm{main}}(H)
&=
\mathcal L_{\mathrm{obs}}^{(v)}
\left(
\operatorname{warp}
\left(
R_v^{\mathrm{main}},H
\right),
I_{v,u}^{\mathrm{ir}};
H
\right),
\\
\ell_v^{\mathrm{mind}}(H)
&=
\operatorname{mean}_{M_v(H),c}
\left|
\operatorname{warp}
\left(
\widehat D_v^{\mathrm{rgb}},H
\right)
-
D_{v,u}^{\mathrm{ir}}
\right|,
\end{aligned}
\label{eq:coarse_registration_cues}
\end{equation}
where $\widehat D_v^{\mathrm{rgb}} =\operatorname{MIND}(R_v^{\mathrm{rgb}})$ and $D_{v,u}^{\mathrm{ir}}=\operatorname{MIND}(I_{v,u}^{\mathrm{ir}})$.
The hat distinguishes the descriptor of the current RGB rendering from $D_v^{\mathrm{rgb}}$, which denotes the descriptor of the observed RGB image during registration initialization.

Let $\mathcal K_H=\{\mathrm{main},\mathrm{mind},\mathrm{ref}\}$.
Together with the fixed-reference cue $\ell_v^{\mathrm{ref}}(H)$ defined in Eq.~\eqref{eq:frc_registration_cue}, the H-block objective is
\begin{equation}
\begin{aligned}
\mathcal L_H^{(v)}
\left(
H_v
\right)
&=
\frac{1}{3}
\sum_{k\in\mathcal K_H}
\frac{
\ell_v^{k}(H_v)
}{
\max
\left(
\operatorname{sg}
\left[
\ell_v^{k}
\left(
H_v^{\mathrm{pre}}
\right)
\right],
\epsilon_H
\right)
}
\\
&\quad+
\frac{\lambda_H}{8}
\left\|
\boldsymbol\delta_v
\right\|_2^2,
\end{aligned}
\label{eq:coarse_homography_objective}
\end{equation}
where $H_v^{\mathrm{pre}}$ is the homography before the current H-block, $\epsilon_H=10^{-8}$, and $\lambda_H=10^{-4}$. Cue-wise normalization makes the update depend on relative improvement rather than absolute loss scale, while the residual regularizer favors local corrections around $H_{0,v}$.
Although block-coordinate optimization removes within-block coupling, it does not prevent co-adaptation across successive blocks.

\subsubsection{Fixed-Reference Consensus Mechanism}
To constrain the remaining inter-block ambiguity, the fixed-reference consensus (FRC) mechanism associates each Gaussian with auxiliary reference IR logits $\mathbf t_i^{\mathrm{ref}}$, with $\mathbf c_i^{\mathrm{ref}}=\sigma(\mathbf t_i^{\mathrm{ref}})$.
These auxiliary logits are kept separate from the main representation $\Theta_R$.

The reference rendering $R_v^{\mathrm{ref}}$ is produced using the detached shared geometry $\operatorname{sg}(\boldsymbol\theta_i)$ and the reference IR appearance $\mathbf c_i^{\mathrm{ref}}$. 
The reference logits are optimized alongside the G-block under the fixed initialized homography:
\begin{equation}
\mathcal L_{\mathrm{ref}}^{(v)}
=
\mathcal L_{\mathrm{obs}}^{(v)}
\left(
\operatorname{warp}
\left(
R_v^{\mathrm{ref}},H_{0,v}
\right),
I_{v,u}^{\mathrm{ir}};
H_{0,v}
\right).
\label{eq:frc_reference_training}
\end{equation}
Because the shared geometry is detached along this path, $\mathcal L_{\mathrm{ref}}^{(v)}$ updates only the reference IR logits and cannot deform the shared geometry to compensate for residual error in $H_{0,v}$.

During the H-block, the detached reference rendering is evaluated under a candidate homography $H$:
\begin{equation}
\ell_v^{\mathrm{ref}}(H)
=
\mathcal L_{\mathrm{obs}}^{(v)}
\left(
\operatorname{warp}
\left(
\operatorname{sg}
\left(
R_v^{\mathrm{ref}}
\right),
H
\right),
I_{v,u}^{\mathrm{ir}};
H
\right).
\label{eq:frc_registration_cue}
\end{equation}
Because the reference IR appearance is learned with supervision coordinates anchored to $H_{0,v}$ rather than the evolving $H_v$, $\ell_v^{\mathrm{ref}}(H)$ provides a complementary registration cue whose appearance supervision does not co-adapt to the current homography trajectory.
After optimizing Eq.~\eqref{eq:coarse_homography_objective}, a proposal $\widetilde H_v$ is rejected if a decrease in $\ell_v^{\mathrm{main}}$ is accompanied by an increase in $\ell_v^{\mathrm{ref}}$ relative to $H_v^{\mathrm{pre}}$; both the homography residual and its optimizer state are then restored. FRC therefore preserves the adaptability of the main IR branch while suppressing homography updates supported only by appearance--registration co-adaptation.

\subsection{Fine Registration--Representation--Fusion Co-Refinement}
Fine co-refinement relaxes shared geometry to accommodate modality-dependent spatial responses because sensor-dependent differences may remain in spatial support and boundary response, while retaining canonical parents as the common support for fusion. Based on this representation, we define view-conditioned pre-projection fusion over the canonical parents, explicitly preserve the cross-modal correspondence required by such primitive-level fusion, and jointly refine representation, fusion, and registration through responsibility-separated optimization.

\subsubsection{Modality-Decoupled Parent--Child Gaussian Representation}
We retain each coarse Gaussian $\mathcal G_i$ as a canonical parent $\mathcal G_i^{p}$ with geometry $\boldsymbol\theta_i=\{\boldsymbol\mu_i,\mathbf s_i,\mathbf q_i,\alpha_i\}$, and associate it with one RGB child and one IR child. Rather than duplicating the parent geometry, the geometry of child $m\in\{\mathrm{rgb},\mathrm{ir}\}$,
$\boldsymbol\theta_i^{m}
=\{\boldsymbol\mu_i^{m},\mathbf s_i^{m},\mathbf q_i,\alpha_i^{m}\}$,
is derived through constrained residuals.

Two independent center residuals would introduce a redundant common translation that overlaps with the parent center. 
We instead parameterize only the relative displacement between the two children and constrain their centers to remain symmetric about the parent:
\begin{equation}
\begin{aligned}
\rho_i
&=
\min
\left(
\mathbf s_i
\right),
\\
\boldsymbol\Delta_i
&=
R(\mathbf q_i)
\left[
\kappa\,
\operatorname{sg}(\rho_i)
\tanh(\mathbf u_i)
\right],
\end{aligned}
\label{eq:modal_relative_displacement}
\end{equation}
where $\mathbf u_i\in\mathbb R^3$ parameterizes the displacement in the parent-local frame, and $\kappa$ bounds each local component relative to the minimum parent scale. 
The child geometries are defined as
\begin{equation}
\begin{aligned}
\boldsymbol\mu_i^{\mathrm{rgb}}
&= \boldsymbol\mu_i - \frac12\boldsymbol\Delta_i,
\mspace{12mu}
\boldsymbol\mu_i^{\mathrm{ir}}
= \boldsymbol\mu_i + \frac12\boldsymbol\Delta_i, \\
\mathbf s_i^{m}
&= \mathbf s_i
\odot
\exp
\Big[
\operatorname{clip}
\big(
\Delta\mathbf s_i^{m},
-\tau_s,
\tau_s
\big)
\Big], \\
\alpha_i^{m}
&= \sigma
\big[
\operatorname{logit}(\alpha_i)
+
\Delta\beta_i^{m}
\big].
\end{aligned}
\label{eq:modal_child_geometry}
\end{equation}
The symmetric center parameterization assigns common motion to the parent and reserves $\boldsymbol\Delta_i$ for cross-modal displacement. Scale and opacity residuals model modality-dependent support and contribution strength.
The two children retain the shared parent rotation, $\mathbf q_i^{\mathrm{rgb}}=\mathbf q_i^{\mathrm{ir}}=\mathbf q_i$, avoiding the ambiguity between independent rotations and anisotropic scales while preserving a common local coordinate frame.

The appearance parameters are inherited from coarse co-refinement. 
The RGB child uses $\mathbf c_{i,v}^{\mathrm{rgb}}=\operatorname{SH}(\mathbf f_i^{\mathrm{SH}},\mathbf d_{i,v}^{\mathrm{rgb}})$, whereas the IR child uses the main IR appearance $\mathbf c_i^{\mathrm{main}}=\sigma(\mathbf t_i^{\mathrm{main}})$.
The auxiliary reference IR appearance remains separate and is used only for registration refinement.

All child residuals are initialized to zero, such that both children initially coincide with their parent and reproduce the coarse representation. 
Fine refinement therefore introduces modality-specific separation only when supported by subsequent supervision.

\subsubsection{View-Conditioned Pre-Projection Fusion}

Fusion is defined over paired modality attributes associated with each canonical parent. For a query camera centered at $\mathbf c_v$, the viewing direction of parent $i$ in its local coordinate frame is
\begin{equation}
\mathbf d_{i,v}^{\mathrm{local}}
=
R(\mathbf q_i)^{\top}
\frac{
\mathbf c_v-\boldsymbol\mu_i
}{
\left\|
\mathbf c_v-\boldsymbol\mu_i
\right\|_2
}.
\label{eq:local_view_direction}
\end{equation}
Each parent is associated with a base logit $b_i$ and a directional coefficient $\mathbf a_i\in\mathbb R^3$. 
Its IR fusion weight is
\begin{equation}
w_{i,v}^{\mathrm{ir}}
=
w_{\min}
+
\left(
w_{\max}-w_{\min}
\right)
\sigma
\left(
b_i+
\mathbf a_i^{\top}
\mathbf d_{i,v}^{\mathrm{local}}
\right).
\label{eq:fusion_weight}
\end{equation}
The base logit captures view-independent modality preference, while the directional term permits view-dependent modulation in the parent-local frame.
The bounded parameterization restricts the admissible mixing range.

The paired RGB and main IR attributes are fused before projection:
\begin{equation}
\mathbf c_{i,v}^{\mathrm{fuse}}
=
\left(
1-w_{i,v}^{\mathrm{ir}}
\right)
\mathbf c_{i,v}^{\mathrm{rgb}}
+
w_{i,v}^{\mathrm{ir}}
\mathbf c_i^{\mathrm{main}}.
\label{eq:preprojection_color}
\end{equation}
The fused attributes are then rendered once using the canonical parent geometry:
\begin{equation}
\hat I_v^{\mathrm{fuse}}
=
\mathcal{R}
\left(
\left\{
\boldsymbol\mu_i,
\mathbf s_i,
\mathbf q_i,
\alpha_i,
\mathbf c_{i,v}^{\mathrm{fuse}}
\right\}_{i=1}^{N}
\right).
\label{eq:single_splat_fusion}
\end{equation}
Unlike blending independently rendered modality images, pre-projection fusion combines paired attributes before a single projection, visibility ordering, and alpha-compositing pass. 
It therefore avoids inconsistent visibility assignments induced by the modality-specific child geometries and uses the parent geometry as the canonical support for fused rendering. 
Because $w_{i,v}^{\mathrm{ir}}$ depends only on the query viewpoint and learned scene parameters, the fusion field can be evaluated at arbitrary viewpoints without an observed RGB--IR pair.

\subsubsection{Cross-Modal Gaussian--Ray Correspondence}
We regularize cross-modal correspondence to support primitive-level fusion, which combines paired RGB and IR appearance attributes and renders the result on the canonical parent geometry.
The parent index fixes their structural pairing, but modality-specific refinement can still lead the paired children to incompatible local scene explanations.
Low reconstruction error in each modality does not ensure that the paired children explain the same local scene region, because different configurations of geometry and appearance can yield similar rendered images.
We associate the paired children with embeddings $\mathbf e_i^{\mathrm{rgb}}, \mathbf e_i^{\mathrm{ir}}\in\mathbb R^{d_e}$ and impose two complementary constraints:
Gaussian-level contrastive learning (GCL) regularizes the paired embeddings before rendering, whereas ray-level contrastive learning (RayCL) constrains their consistency through projection and alpha compositing.

For a sampled parent set $\mathcal B$, the normalized child embeddings and the cross-modal similarity matrix are defined as
$\bar{\mathbf e}_i^{m}
=\mathbf e_i^{m}/\|\mathbf e_i^{m}\|_2$,
where $m\in\{\mathrm{rgb},\mathrm{ir}\}$, and
$S_{ij}
=\langle\bar{\mathbf e}_i^{\mathrm{rgb}},
\bar{\mathbf e}_j^{\mathrm{ir}}\rangle/\tau_g$
for $i,j\in\mathcal B$.
Children derived from the same parent form the diagonal positive pairs, while children from different parents serve as negatives.
The Gaussian-level contrastive objective is defined as
\begin{equation}
\mathcal L_{\mathrm{GCL}}
=
\frac{1}{2}
\left[
\operatorname{CE}
\left(
S,\operatorname{diag}
\right)
+
\operatorname{CE}
\left(
S^{\top},\operatorname{diag}
\right)
\right],
\label{eq:gcl}
\end{equation}
where $\operatorname{diag}$ denotes the diagonal parent pairing.
GCL regularizes the predefined pairing in embedding space without directly constraining child geometry.

To couple these associations to image formation, we treat the normalized embeddings as Gaussian feature colors and render modality-specific feature maps using the corresponding child geometries:
\begin{equation}
\mathbf c_i^{\mathrm{feat},m}
=
\frac{1}{2}
\left(
\bar{\mathbf e}_i^{m}
+
\mathbf 1
\right),
\qquad
m\in\{\mathrm{rgb},\mathrm{ir}\}.
\label{eq:gaussian_feature_color}
\end{equation}
\begin{equation}
\widetilde F_v^{m}
=
\mathcal R
\left(
\mathcal G^{m};
\left\{
\mathbf c_i^{\mathrm{feat},m}
\right\}_{i=1}^{N}
\right).
\label{eq:rendered_feature_map}
\end{equation}
Mapping the rendered values back to the embedding domain and normalizing along the feature dimension gives
$F_v^{m}
=
\operatorname{norm}
\left(
2\widetilde F_v^{m}
-
\mathbf 1
\right)$.
Their correspondence is enforced in the image plane on rays jointly supported by the two modality-specific renderings:
\begin{equation}
\mathcal L_{\mathrm{ray}}^{\mathrm{cross}}
=
\operatorname{InfoNCE}
\left(
F_v^{\mathrm{rgb}},
F_v^{\mathrm{ir}};
M_v^{\mathrm{ray}}
\right),
\label{eq:ray_cross}
\end{equation}
where $M_v^{\mathrm{ray}}$ is obtained as the intersection of the alpha supports
of the RGB and IR feature renderings in the canonical camera domain.

Cross-modal agreement between the two rendered branches alone could still yield an internally consistent feature field that is weakly anchored to the received thermal observation. 
We therefore additionally evaluate the IR feature rendering in the original
IR observation domain:
\begin{equation}
\mathcal L_{\mathrm{ray}}^{\mathrm{obs}}
=
\operatorname{InfoNCE}
\left(
\operatorname{warp}
\left(
F_v^{\mathrm{ir}},
\operatorname{sg}(H_v)
\right),
\bar I_{v,u}^{\mathrm{ir}};
M_v^{\mathrm{obs}}
\right),
\label{eq:ray_obs}
\end{equation}
where $\bar I_{v,u}^{\mathrm{ir}}=\operatorname{norm}(2I_{v,u}^{\mathrm{ir}}-\mathbf 1)$ denotes the normalized IR observation, and $M_v^{\mathrm{obs}}$ denotes the valid pixels where the warped IR feature rendering overlaps with the observed IR image. 
The ray-level objective is
\begin{equation}
\mathcal L_{\mathrm{RayCL}}
=
\frac{1}{2}
\left(
\mathcal L_{\mathrm{ray}}^{\mathrm{cross}}
+
\mathcal L_{\mathrm{ray}}^{\mathrm{obs}}
\right).
\label{eq:raycl}
\end{equation}
Unlike GCL, RayCL propagates through child projection and alpha compositing and therefore also constrains the modality-specific geometry responsible for the rendered correspondence. 
The current homography is detached throughout this path, preventing correspondence loss from being reduced through simultaneous registration adaptation.

The overall correspondence objective is
\begin{equation}
\mathcal L_{\mathrm{corr}}
=
\mathcal L_{\mathrm{GCL}}
+
\lambda_{\mathrm{ray}}
\mathcal L_{\mathrm{RayCL}}.
\label{eq:correspondence_loss}
\end{equation}
These losses support pre-projection fusion by regularizing its underlying scene representation. GCL constrains the paired embeddings, while RayCL couples their rendered consistency to the shared parent geometry and child residuals.
The fusion field combines the learned RGB and IR appearance attributes on the canonical parents, without using the embeddings as fusion inputs.

\subsubsection{Reliability-Constrained Tri-Task Co-Refinement}
Fine co-refinement retains the G/H alternation introduced in \ref{sec:H/G}. The representation parameters $\Theta_R$ are extended to include the child geometric residuals and correspondence embeddings, while $\Theta_F=\{b_i,\mathbf a_i\}_{i=1}^{N}$ denotes the fusion parameters. 
The two responsibility-separated updates are
\begin{equation}
\begin{aligned}
\left(
\Theta_R^{+},
\Theta_F^{+}
\right)
&\leftarrow
\operatorname{Opt}_{G}
\left(
\Theta_R,
\Theta_F;
\operatorname{sg}(\mathbf H)
\right),
&&
\text{G-block},
\\
\boldsymbol\delta_v^{+}
&\leftarrow
\operatorname{Opt}_{H}
\left(
\boldsymbol\delta_v;
\operatorname{sg}(\Theta_R)
\right),
&&
\text{H-block}.
\end{aligned}
\label{eq:fine_alternating_refinement}
\end{equation}
The G-block jointly updates representation and fusion without changing the observation coordinates, whereas the H-block updates registration against frozen scene evidence. 
The fixed-reference IR branch remains separate from $\Theta_R$ and is retained only as registration evidence.

\paragraph{Scene-Side Representation and Fusion Refinement}
In the fine G-block, RGB and IR children are optimized using the reconstruction
losses defined in Eq. \eqref{eq:coarse_main_objective}, with the corresponding child geometries used for rendering.
The fixed view-wise reliability weights are retained.

For training view $v$, the observed IR image is aligned to the canonical RGB
domain using the inverse of the detached current homography:
\begin{equation}
I_{v,a}^{\mathrm{ir}}
=
\operatorname{warp}
\left(
I_{v,u}^{\mathrm{ir}},
\operatorname{sg}
\left(
H_v^{-1}
\right)
\right),
\label{eq:aligned_ir_observation}
\end{equation}
where subscript $a$ denotes the homography-aligned IR observation, and
$M_v^a=M_v(H_v^{-1})$ denotes the corresponding valid support.
Fusion supervision combines intensity and gradient losses
over the aligned valid support:
\begin{equation}
\begin{aligned}
\mathcal L_{\mathrm{int}}^{(v)}
&=
\operatorname{mean}_{M_v^a,c}
\left|
\hat I_v^{\mathrm{fuse}}
-
\max\!\left(I_v^{\mathrm{rgb}},I_{v,a}^{\mathrm{ir}}\right)
\right|,
\\
\mathcal L_{\mathrm{grad}}^{(v)}
&=
\operatorname{mean}_{M_v^a}
\Big\|
\big|\nabla_{\mathrm S}\hat I_v^{\mathrm{fuse}}\big|
\\
&\quad -
\max\!\left(
\big|\nabla_{\mathrm S}I_v^{\mathrm{rgb}}\big|,
\big|\nabla_{\mathrm S}I_{v,a}^{\mathrm{ir}}\big|
\right)
\Big\|_1,
\\
\mathcal L_{\mathrm{fuse}}^{(v)}
&=
\mathcal L_{\mathrm{int}}^{(v)}
+
\lambda_{\mathrm{grad}}
\mathcal L_{\mathrm{grad}}^{(v)}.
\end{aligned}
\label{eq:fusion_reconstruction}
\end{equation}
Here, $\lambda_{\mathrm{grad}}$ weights the gradient term,
$\nabla_{\mathrm S}$ is the previously defined Sobel operator,
and absolute values and maxima are applied elementwise.

Because the directional coefficients are learned from finitely many training views, we use $\mathcal L_{\mathrm{view}}=(3N)^{-1}\sum_{i=1}^{N}\|\mathbf a_i\|_2^2$ to favor a view-independent modality preference unless directional modulation is supported by fusion supervision. The fusion objective is $\mathcal L_F^{(v)}=\lambda_f\mathcal L_{\mathrm{fuse}}^{(v)}+\lambda_{\mathrm{view}}\mathcal L_{\mathrm{view}}$.

After averaging view-dependent terms over the sampled training views, the fine G-block objective is
\begin{equation}
\mathcal L_G
=
\mathcal L_{\mathrm{rec}}
+
\lambda_{\mathrm{corr}}
\mathcal L_{\mathrm{corr}}
+
\mathcal L_F
+
\lambda_{\mathrm{reg}}
\mathcal L_{\mathrm{reg}},
\label{eq:fine_scene_objective}
\end{equation}
where $\mathcal L_{\mathrm{reg}}$ penalizes excessive child displacement, scale, and opacity residuals, preventing unsupported deviation from the canonical parents.

\paragraph{RGB-Protected Geometry Update}

Although modality-specific residuals absorb sensor-dependent variation, both children remain differentiable functions of the parent geometry. 
RGB and main IR reconstruction losses can therefore produce conflicting gradients on the shared parent parameters. Because the parent geometry and camera poses are anchored by the RGB reconstruction, we retain the RGB gradient as the reference direction and remove only the conflicting component of the main IR gradient.

Let $\mathbf g_i^{\mathrm{rgb}}$ and $\mathbf g_i^{\mathrm{ir}}$ denote the RGB and weighted main IR reconstruction gradients, respectively, on the parent center, scale, and opacity. When
$\langle
\mathbf g_i^{\mathrm{rgb}},
\mathbf g_i^{\mathrm{ir}}
\rangle<0$, the main IR contribution is replaced by~\cite{yu2020gradient}
\begin{equation}
\widetilde{\mathbf g}_i^{\mathrm{ir}}
=
\mathbf g_i^{\mathrm{ir}}
-
\frac{
\left\langle
\mathbf g_i^{\mathrm{ir}},
\mathbf g_i^{\mathrm{rgb}}
\right\rangle
}{
\left\|
\mathbf g_i^{\mathrm{rgb}}
\right\|_2^2
+
\epsilon_g
}
\mathbf g_i^{\mathrm{rgb}}.
\label{eq:rgb_protected_gradient}
\end{equation}
Otherwise,
$\widetilde{\mathbf g}_i^{\mathrm{ir}}
=\mathbf g_i^{\mathrm{ir}}$.
This correction modifies only the conflicting main IR reconstruction contribution; gradients from RGB reconstruction, fusion, correspondence, and regularization remain unchanged.
It  therefore  prevents modality-specific disagreement from distorting the canonical parent.

\paragraph{Reliability-Constrained Registration Update}
The fine H-block inherits the homography parameterization and normalized main-IR, MIND, and fixed-reference objective from coarse co-refinement.
After modality-specific child geometry is introduced,
however, an image residual no longer provides unambiguous evidence of misregistration: it may also arise from legitimate sensor-dependent spatial variation or rendering approximation.
Registration is therefore updated only for views with substantial residual evidence.

For each view, we define the appearance and structural evidence as
$r_v^{\mathrm{app}}
=\max\{
\ell_v^{\mathrm{main}}(H_v),
\ell_v^{\mathrm{ref}}(H_v)
\}$ and
$r_v^{\mathrm{str}}
=\ell_v^{\mathrm{mind}}(H_v)$.
Let $\pi_v^{\mathrm{app}}$ and $\pi_v^{\mathrm{str}}$ denote their scene-relative percentile ranks, with larger ranks indicating larger residuals.
The active view set is
\begin{equation}
\mathcal V_H
=
\left\{
v
\;\middle|\;
\pi_v^{\mathrm{app}}
\geq
\tau_{\mathrm{app}}^{r}
\;\lor\;
\pi_v^{\mathrm{str}}
\geq
\tau_{\mathrm{str}}^{r}
\right\}.
\label{eq:fine_registration_active_set}
\end{equation}
A view is therefore left unchanged only when neither evidence family indicates substantial remaining inconsistency.

For each $v\in\mathcal V_H$, we optimize the inherited three-cue objective in Eq.~\eqref{eq:coarse_homography_objective} against the frozen fine representation. 
Let $\widetilde H_v$ denote the resulting proposal. 
The accepted update is
\begin{equation}
H_v^{+}
=
\begin{cases}
\widetilde H_v,
&
\Delta_v>0
\ \land\
\widetilde H_v
\in
\mathcal H_v^{\mathrm{valid}},
\\
H_v,
&
\text{otherwise},
\end{cases}
\label{eq:fine_registration_acceptance}
\end{equation}
where $\Delta_v$ is the aggregate relative decrease of the frozen main-IR, MIND, and fixed-reference cues, and $\mathcal H_v^{\mathrm{valid}}$ denotes the trusted set imposed by the geometric validity and bounded-update constraints. 
Rejected proposals restore both the homography residual and its optimizer state. 
This conservative update retains the current correspondence unless multiple frozen cues support a geometrically plausible refinement.

\section{Experiments}
\label{sec:experiment}
This section describes the datasets and experimental settings, followed by quantitative and qualitative comparisons with representative infrared--visible fusion baselines and ablation studies of the proposed framework.

\subsection{Datasets and Baselines}
We evaluate our method on the expanded RGBT-Scenes~\cite{lu2024thermalgaussian} dataset, which comprises 14 indoor and outdoor scenes spanning industrial facilities, open environments, buildings, and everyday objects at different spatial scales.
We additionally use 4 scenes from ThermoScenes~\cite{hassan2025thermonerf}, covering building facades captured under different seasonal and temperature conditions.
Each scene comprises tens to more than one hundred aligned RGB--IR image pairs acquired from different viewpoints, providing either $360^{\circ}$ or forward-facing observation coverage. 
The images of each scene are divided into train- and test- sets.
To emulate and deliberately amplify the cross-modal misregistration encountered in real-world dual-spectrum systems, we independently perturb the IR image of each observed pair using a random projective transformation in a controlled manner.
Specifically, the transformation involves an isotropic scaling factor of $[0.8,1.2]$, perspective perturbations of $[-0.15,0.15]$ along both spatial axes, and an in-plane rotation of $[-15^{\circ},15^{\circ}]$.
This protocol enables a systematic evaluation of robustness under challenging cross-modal geometric discrepancies.

We compare our method with both pairwise IVF methods and scene-level cascade baselines. For misaligned RGB--IR inputs, the pairwise methods follow the two prevailing strategies of joint registration and fusion and registration followed by fusion. The former includes IMF~\cite{wang2024improving}, MulFS-CAP~\cite{li2025mulfs}, C2RF~\cite{tang2025c2rf}, HMMF~\cite{xiao2026hybrid}, and FusionRegister~\cite{bian2026fusionregister} equipped with MMDRFuse~\cite{deng2024mmdrfuse} as its fusion backbone. For the latter, XoFTR~\cite{tuzcuouglu2024xoftr} serves as a cross-modal matcher whose predicted correspondences are converted into a global homography using RANSAC~\cite{fischler1981random}, whereas CRFT~\cite{liu2026crft} directly estimates a dense cross-modal deformation field. Each registered RGB--IR pair is subsequently fused using CDMFusion~\cite{yang2025cdmfusion} or Masked-DiFuser~\cite{tang2025mask}. To contextualize the effect of residual misalignment, we additionally evaluate these two fusion methods on reference-aligned RGB--IR pairs as aligned-input pairwise references. 
We further construct two types of scene-level cascade baselines. For fusion-then-reconstruction, the joint registration and fusion method HMMF and the registration and fusion pipeline comprising CRFT followed by Masked-DiFuser are separately applied to all training pairs, after which a standard 3DGS~\cite{kerbl20233d} is optimized from the resulting fused images. For reconstruction-then-fusion, CRFT first aligns the training RGB and IR observations, OMMG~\cite{lu2026thermalgaussian++} reconstructs and renders the two modalities at each query viewpoint, and Masked-DiFuser subsequently fuses the rendered RGB and IR pair. These cascades respectively perform pairwise fusion before scene reconstruction and image-space fusion after dual-modal rendering, providing task-matched scene-level comparisons with our jointly optimized pre-projection fusion framework.

\subsection{Implementation Details}

\label{sec:implementation_details} 
We use Adam~\cite{kingma2014adam} for all optimization stages and all experiments are conducted on an NVIDIA RTX 3090 GPU.
\par\noindent\textbf{Registration Initialization.}\enspace
For each view, we sample $K=512$ homography candidates and refine the selected candidate over a four-level pyramid with longer-side resolutions $\{80,160,320,640\}$. Each level is optimized for 120 iterations with a learning rate of $10^{-2}$. We set $\lambda_{\mathrm{ngf}}=0.2$, $\lambda_{\eta}=10^{-3}$, and $\tau_{\mathrm{init}}=0.6$. 
\par\noindent\textbf{Coarse Co-Refinement.}\enspace
Coarse co-refinement is performed at half the input resolution using 1,632 G-updates and 360 view-specific H-updates. It starts with 192 G-updates, alternates 32-update G-blocks with 8-update H-blocks, and ends with 128 appearance-only G-updates. The learning rates are $2.5\times10^{-3}$ for the homography residuals and $2\times10^{-2}$ for the main and reference IR logits. We use $\lambda_{\mathrm{rgb}}=0.8$, $\lambda_{\mathrm{ssim}}=0.2$, $\epsilon_c=10^{-6}$, $\lambda_{\nabla}=0.2$, $\tau_o=0.45$, $\epsilon_H=10^{-8}$, and $\lambda_H^{\mathrm{coarse}}=10^{-4}$. The gradient norm is clipped to 5. 
\par\noindent\textbf{Fine Co-Refinement.}\enspace
Fine co-refinement is conducted at the full input resolution using 1,500 G-updates, including 50 initial and 50 terminal G-updates. It alternates 10 G-updates and 8 H-updates. The learning rates for the parent center, SH DC coefficients, higher-order SH coefficients, opacity, scale, rotation, and main IR logits are respectively $10^{-5}$, $10^{-3}$, $5\times10^{-5}$, $5\times10^{-4}$, $2\times10^{-4}$, $10^{-4}$, and $10^{-3}$. The child geometric residuals use $5\times10^{-4}$, whereas the embeddings and fusion parameters use $10^{-3}$; the reference IR logits are frozen. We set $\kappa=0.35$, $\tau_s=1.5$, $d_e=3$, $[w_{\min},w_{\max}]=[0.05,0.95]$, and $\tau_g=0.12$. The correspondence objective uses $\lambda_{\mathrm{ray}}=2$ and $\lambda_{\mathrm{corr}}=0.005$. For fusion supervision, we use $\epsilon_f=10^{-6}$, $\lambda_f=0.25$, $\lambda_{\mathrm{grad}}=2$ and $\lambda_{\mathrm{view}}=5\times10^{-4}$. The effective weights of the child-displacement, scale-residual, and opacity-residual penalties are $0.01$, $0.002$, and $0.002$, respectively. The fine H-block uses a learning rate of $2\times10^{-3}$, and gradient clipping at 2. Finally, we set $\tau_{\mathrm{app}}^{r}=0.45$ and $\tau_{\mathrm{str}}^{r}=0.25$, with global and block-wise residual bounds of $0.02$ and $0.008$, respectively.

\begin{table*}[t]
\centering
\caption{
Quantitative fusion comparison at pair-available held-out test viewpoints.
Pairwise methods receive the residually misaligned RGB--IR pair at the test viewpoint, whereas scene-level methods synthesize the fused image using only the query viewpoint.
Fusion methods receiving reference-aligned pairs are reported separately as contextual references.
FTR and RTF denote fusion-then-reconstruction and reconstruction-then-fusion, respectively.
Within each dataset, the best and second-best cells are highlighted in light green and light blue for pairwise IVF methods and in light red and light yellow for scene-level methods.
Aligned-input references are excluded from ranking.
}
\label{tab:free_view_fusion}
\setlength{\tabcolsep}{2.0pt}
\renewcommand{\arraystretch}{1.06}
\footnotesize
\resizebox{\textwidth}{!}{%
\begin{tabular}{l*{9}{c}}
\toprule
\multicolumn{10}{c}{\textbf{RGBT-Scenes}}\\
\midrule
Method
& EN$\uparrow$
& $\mathrm{VIF}^{\mathrm{rgb}}\uparrow$
& $\mathrm{VIF}^{\mathrm{ir}}\uparrow$
& SCD$\uparrow$
& $Q_{abf}^{\mathrm{rgb}}\uparrow$
& $Q_{abf}^{\mathrm{ir}}\uparrow$
& $\mathrm{SSIM}_d\uparrow$
& Opt.\,(s)$\downarrow$
& FPS$\uparrow$
\\
\midrule

\multicolumn{10}{l}{
\textit{Pairwise methods}
}\\

IMF
& $7.204 \pm 0.325$
& $0.372 \pm 0.119$
& $0.126 \pm 0.053$
& $1.020 \pm 0.418$
& $0.316 \pm 0.130$
& $0.043 \pm 0.016$
& $0.511 \pm 0.068$
& --
& $12.326 \pm 0.093$
\\

MulFS-CAP
& $6.864 \pm 0.414$
& $0.183 \pm 0.057$
& $0.092 \pm 0.029$
& $0.821 \pm 0.430$
& $0.276 \pm 0.094$
& $0.047 \pm 0.021$
& $0.486 \pm 0.071$
& --
& $0.058 \pm 0.001$
\\

C2RF
& $7.410 \pm 0.215$
& $0.451 \pm 0.088$
& $0.141 \pm 0.049$
& $1.160 \pm 0.345$
& $0.436 \pm 0.108$
& \pairfirst{$0.053 \pm 0.023$}
& $0.539 \pm 0.072$
& --
& \pairsecond{$12.549 \pm 0.132$}
\\

HMMF
& $7.291 \pm 0.400$
& \pairfirst{$0.853 \pm 0.071$}
& $0.104 \pm 0.044$
& $0.967 \pm 0.353$
& \pairfirst{$0.655 \pm 0.105$}
& \pairsecond{$0.050 \pm 0.024$}
& \pairfirst{$0.586 \pm 0.070$}
& --
& $7.875 \pm 0.064$
\\

FusionRegister(MMDRFuse)
& $7.116 \pm 0.320$
& $0.447 \pm 0.118$
& $0.116 \pm 0.047$
& $1.015 \pm 0.455$
& $0.379 \pm 0.137$
& $0.049 \pm 0.022$
& \pairsecond{$0.552 \pm 0.076$}
& --
& \pairfirst{$14.087 \pm 0.172$}
\\

XoFTR+CDMFusion
& $7.446 \pm 0.267$
& $0.487 \pm 0.221$
& \pairfirst{$0.162 \pm 0.060$}
& $1.088 \pm 0.262$
& $0.432 \pm 0.175$
& $0.044 \pm 0.020$
& $0.493 \pm 0.065$
& --
& $0.385 \pm 0.079$
\\

XoFTR+Masked-DiFuser
& \pairsecond{$7.561 \pm 0.141$}
& \pairsecond{$0.564 \pm 0.128$}
& \pairsecond{$0.150 \pm 0.059$}
& \pairsecond{$1.245 \pm 0.284$}
& \pairsecond{$0.460 \pm 0.125$}
& $0.040 \pm 0.016$
& $0.498 \pm 0.057$
& --
& $0.130 \pm 0.001$
\\

CRFT+CDMFusion
& $7.470 \pm 0.266$
& $0.440 \pm 0.225$
& $0.128 \pm 0.054$
& $1.083 \pm 0.286$
& $0.404 \pm 0.174$
& $0.034 \pm 0.014$
& $0.478 \pm 0.068$
& --
& $0.444 \pm 0.099$
\\

CRFT+Masked-DiFuser
& \pairfirst{$7.607 \pm 0.141$}
& $0.531 \pm 0.103$
& $0.126 \pm 0.053$
& \pairfirst{$1.261 \pm 0.305$}
& $0.452 \pm 0.107$
& $0.034 \pm 0.015$
& $0.487 \pm 0.053$
& --
& $0.122 \pm 0.016$
\\

\midrule
\multicolumn{10}{l}{
\textit{Aligned-input references}
}\\

CDMFusion (aligned)
& $7.460 \pm 0.250$
& $0.421 \pm 0.229$
& $0.442 \pm 0.182$
& $1.521 \pm 0.174$
& $0.399 \pm 0.172$
& $0.105 \pm 0.054$
& $0.516 \pm 0.058$
& --
& $0.483 \pm 0.093$
\\

Masked-DiFuser (aligned)
& $7.595 \pm 0.138$
& $0.491 \pm 0.140$
& $0.341 \pm 0.099$
& $1.843 \pm 0.064$
& $0.428 \pm 0.116$
& $0.081 \pm 0.033$
& $0.518 \pm 0.058$
& --
& $0.135 \pm 0.000$
\\

\midrule
\multicolumn{10}{l}{
\textit{Scene-level methods}
}\\

FTR: HMMF$\rightarrow$3DGS
& $7.310 \pm 0.358$
& \scenefirst{$0.390 \pm 0.073$}
& $0.102 \pm 0.033$
& $0.768 \pm 0.380$
& \scenefirst{$0.323 \pm 0.084$}
& \scenesecond{$0.053 \pm 0.026$}
& \scenesecond{$0.527 \pm 0.070$}
& \scenefirst{$687.645 \pm 330.442$}
& \scenefirst{$272.895 \pm 63.869$}
\\

FTR: CRFT+Masked-DiFuser$\rightarrow$3DGS
& \scenesecond{$7.635 \pm 0.133$}
& $0.267 \pm 0.050$
& $0.116 \pm 0.040$
& $1.054 \pm 0.342$
& $0.244 \pm 0.072$
& $0.044 \pm 0.019$
& $0.452 \pm 0.049$
& $1366.035 \pm 584.991$
& $248.863 \pm 49.646$
\\

RTF: CRFT+OMMG$\rightarrow$Masked-DiFuser
& \scenefirst{$7.648 \pm 0.137$}
& \scenesecond{$0.329 \pm 0.051$}
& \scenesecond{$0.126 \pm 0.044$}
& \scenesecond{$1.289 \pm 0.251$}
& \scenesecond{$0.287 \pm 0.077$}
& $0.046 \pm 0.020$
& $0.478 \pm 0.054$
& \scenesecond{$1110.225 \pm 123.157$}
& $0.147 \pm 0.000$
\\

\textbf{Beyond3D (Ours)}
& $7.566 \pm 0.347$
& $0.262 \pm 0.082$
& \scenefirst{$0.196 \pm 0.054$}
& \scenefirst{$1.433 \pm 0.213$}
& $0.197 \pm 0.076$
& \scenefirst{$0.064 \pm 0.032$}
& \scenefirst{$0.588 \pm 0.080$}
& $1152.189 \pm 117.491$
& \scenesecond{$257.567 \pm 25.946$}
\\

\specialrule{0.8pt}{3pt}{2pt}
\multicolumn{10}{c}{\textbf{ThermoScenes}}\\
\midrule
Method
& EN$\uparrow$
& $\mathrm{VIF}^{\mathrm{rgb}}\uparrow$
& $\mathrm{VIF}^{\mathrm{ir}}\uparrow$
& SCD$\uparrow$
& $Q_{abf}^{\mathrm{rgb}}\uparrow$
& $Q_{abf}^{\mathrm{ir}}\uparrow$
& $\mathrm{SSIM}_d\uparrow$
& Opt.\,(s)$\downarrow$
& FPS$\uparrow$
\\
\midrule

\multicolumn{10}{l}{
\textit{Pairwise methods}
}\\

IMF
& $6.026 \pm 0.619$
& $0.316 \pm 0.141$
& $0.117 \pm 0.027$
& $1.230 \pm 0.359$
& $0.206 \pm 0.163$
& $0.097 \pm 0.067$
& $0.576 \pm 0.093$
& --
& \pairsecond{$13.473 \pm 1.230$}
\\

MulFS-CAP
& $5.957 \pm 0.510$
& $0.177 \pm 0.037$
& $0.105 \pm 0.042$
& \pairsecond{$1.339 \pm 0.307$}
& $0.233 \pm 0.127$
& $0.111 \pm 0.066$
& $0.572 \pm 0.075$
& --
& $0.066 \pm 0.009$
\\

C2RF
& $6.186 \pm 0.439$
& $0.263 \pm 0.028$
& \pairfirst{$0.171 \pm 0.037$}
& $1.159 \pm 0.332$
& $0.219 \pm 0.101$
& \pairsecond{$0.117 \pm 0.076$}
& $0.573 \pm 0.065$
& --
& $12.888 \pm 0.400$
\\

HMMF
& $6.254 \pm 0.626$
& \pairsecond{$0.936 \pm 0.174$}
& $0.086 \pm 0.024$
& \pairfirst{$1.437 \pm 0.480$}
& \pairfirst{$0.521 \pm 0.258$}
& \pairfirst{$0.120 \pm 0.083$}
& \pairsecond{$0.577 \pm 0.089$}
& --
& $8.790 \pm 1.092$
\\

FusionRegister(MMDRFuse)
& $5.833 \pm 0.398$
& $0.340 \pm 0.032$
& $0.093 \pm 0.034$
& $1.199 \pm 0.440$
& $0.247 \pm 0.117$
& $0.109 \pm 0.066$
& \pairfirst{$0.597 \pm 0.088$}
& --
& \pairfirst{$15.388 \pm 1.506$}
\\

XoFTR+CDMFusion
& $5.877 \pm 0.857$
& $0.371 \pm 0.184$
& \pairsecond{$0.135 \pm 0.076$}
& $0.737 \pm 0.295$
& $0.287 \pm 0.197$
& $0.047 \pm 0.014$
& $0.504 \pm 0.092$
& --
& $0.468 \pm 0.099$
\\

XoFTR+Masked-DiFuser
& \pairsecond{$6.801 \pm 0.650$}
& \pairfirst{$0.996 \pm 0.545$}
& $0.108 \pm 0.046$
& $0.889 \pm 0.357$
& $0.403 \pm 0.228$
& $0.048 \pm 0.025$
& $0.453 \pm 0.123$
& --
& $0.158 \pm 0.031$
\\

CRFT+CDMFusion
& $6.044 \pm 0.470$
& $0.201 \pm 0.063$
& $0.119 \pm 0.019$
& $0.892 \pm 0.348$
& $0.200 \pm 0.123$
& $0.048 \pm 0.030$
& $0.523 \pm 0.061$
& --
& $0.598 \pm 0.095$
\\

CRFT+Masked-DiFuser
& \pairfirst{$7.103 \pm 0.593$}
& $0.870 \pm 0.278$
& $0.124 \pm 0.031$
& $1.135 \pm 0.368$
& \pairsecond{$0.412 \pm 0.226$}
& $0.046 \pm 0.040$
& $0.498 \pm 0.074$
& --
& $0.121 \pm 0.029$
\\

\midrule
\multicolumn{10}{l}{
\textit{Aligned-input references}
}\\

CDMFusion (aligned)
& $5.953 \pm 0.113$
& $0.213 \pm 0.087$
& $0.620 \pm 0.365$
& $1.265 \pm 0.602$
& $0.255 \pm 0.138$
& $0.184 \pm 0.147$
& $0.595 \pm 0.062$
& --
& $0.498 \pm 0.088$
\\

Masked-DiFuser (aligned)
& $7.341 \pm 0.407$
& $0.627 \pm 0.136$
& $0.770 \pm 0.464$
& $1.456 \pm 0.473$
& $0.389 \pm 0.215$
& $0.116 \pm 0.084$
& $0.522 \pm 0.064$
& --
& $0.134 \pm 0.027$
\\

\midrule
\multicolumn{10}{l}{
\textit{Scene-level methods}
}\\

FTR: HMMF$\rightarrow$3DGS
& $6.706 \pm 0.431$
& \scenefirst{$0.341 \pm 0.110$}
& $0.093 \pm 0.014$
& $1.273 \pm 0.325$
& \scenefirst{$0.206 \pm 0.097$}
& \scenesecond{$0.107 \pm 0.071$}
& \scenesecond{$0.544 \pm 0.051$}
& \scenefirst{$871.155 \pm 375.800$}
& \scenefirst{$285.369 \pm 116.381$}
\\

FTR: CRFT+Masked-DiFuser$\rightarrow$3DGS
& \scenefirst{$7.394 \pm 0.321$}
& $0.274 \pm 0.074$
& \scenesecond{$0.119 \pm 0.022$}
& $1.190 \pm 0.389$
& $0.185 \pm 0.092$
& $0.089 \pm 0.078$
& $0.447 \pm 0.086$
& $1916.710 \pm 843.236$
& $239.196 \pm 72.643$
\\

RTF: CRFT+OMMG$\rightarrow$Masked-DiFuser
& \scenesecond{$7.153 \pm 0.698$}
& \scenesecond{$0.338 \pm 0.101$}
& $0.116 \pm 0.033$
& \scenesecond{$1.298 \pm 0.276$}
& \scenesecond{$0.204 \pm 0.096$}
& $0.091 \pm 0.078$
& $0.457 \pm 0.079$
& $1446.340 \pm 507.880$
& $0.156 \pm 0.022$
\\

\textbf{Beyond3D (Ours)}
& $6.947 \pm 0.324$
& $0.220 \pm 0.070$
& \scenefirst{$0.256 \pm 0.067$}
& \scenefirst{$1.463 \pm 0.285$}
& $0.155 \pm 0.054$
& \scenefirst{$0.124 \pm 0.037$}
& \scenefirst{$0.694 \pm 0.026$}
& \scenesecond{$1243.579 \pm 127.194$}
& \scenesecond{$273.881 \pm 19.469$}
\\

\bottomrule
\end{tabular}%
}
\end{table*}

\subsection{Free-View Infrared--Visible Fusion}
We first assess fusion quality at held-out test views through quantitative and qualitative comparisons, and then examine fused rendering along selected camera trajectories.
\subsubsection{Quantitative Comparison at Test Views}
We evaluate fusion quality at held-out test viewpoints whose reference RGB and IR observations are excluded from scene optimization.
Pairwise IVF methods directly process the test-view image pair while scene-level methods instead synthesize the fused image from the query viewpoint alone, with the corresponding test images used exclusively as evaluation references.
The scene-level baselines therefore constitute the task-matched comparison for free-view IVF, while pairwise methods contextualize the quality attainable with direct access to test-view observations.

We employ commonly used metrics in infrared--visible image fusion, including entropy (EN), visual information fidelity (VIF)~\cite{sheikh2006image}, the sum of correlations of differences (SCD)~\cite{aslantas2015new}, $Q_{abf}$~\cite{xydeas2000objective}, and dual-reference structural similarity $\mathrm{SSIM}_d$.
EN measures the information content of the fused image.
VIF quantifies information transferred from the source modalities, with $\mathrm{VIF}^{\mathrm{rgb}}$ and $\mathrm{VIF}^{\mathrm{ir}}$ reporting their respective contributions.
SCD measures the preservation of complementary source information.
$Q_{abf}$ evaluates edge preservation, with $Q_{abf}^{\mathrm{rgb}}$ and $Q_{abf}^{\mathrm{ir}}$ indicating modality-specific contributions.
$\mathrm{SSIM}_d$ is computed as the average SSIM~\cite{wang2004image} between the fused image and the reference RGB and aligned IR images.
We report the one-time scene optimization time for scene-level methods and the inference speed in frames per second for all methods.

As shown in Table~\ref{tab:free_view_fusion}, three observations can be made.
First, among methods that do not receive reference-aligned pairs, Beyond3D provides the strongest thermal information retention, achieving the highest $\mathrm{VIF}^{\mathrm{ir}}$ and $Q_{abf}^{\mathrm{ir}}$ on both datasets.
Most pairwise methods operating on misaligned inputs and all competing scene-level cascades exhibit substantially larger RGB than IR contributions, suggesting that their fused outputs are dominated by visible information and assign comparatively limited emphasis to thermal structures.
Beyond3D narrows this modality gap by substantially increasing the thermal contribution, although its RGB-specific scores are lower than those of several pairwise methods.
It therefore achieves a substantially more balanced RGB--IR information distribution rather than simultaneously maximizing both modality-specific terms, while preserving competitive overall structural similarity.
Second, registration accuracy has a direct impact on fusion quality.
The aligned-input CDMFusion and Masked-Difuser references substantially outperform their XoFTR- and CRFT-based pipelines in $\mathrm{VIF}^{\mathrm{ir}}$, SCD, and $Q_{abf}^{\mathrm{ir}}$ on both datasets, and generally outperform the joint registration and fusion methods in thermal information retention.
These gaps show that residual geometric errors propagate into the fusion stage and suppress both thermal transfer and cross-modal complementarity.
Third, Beyond3D consistently surpasses all scene-level cascades in $\mathrm{VIF}^{\mathrm{ir}}$, SCD, $Q_{abf}^{\mathrm{ir}}$, and $\mathrm{SSIM}_d$.
Relative to the strongest cascade result for each metric, the improvements reach $55.6\%$, $11.2\%$, $20.8\%$, and $11.6\%$ on RGBT-Scenes, and $115.1\%$, $12.7\%$, $15.9\%$, and $27.6\%$ on ThermoScenes.
Despite receiving no target-view image pair, Beyond3D also exceeds the aligned-input references in $\mathrm{SSIM}_d$ on both datasets and attains the highest overall SCD on ThermoScenes.
These results demonstrate competitive overall fusion quality relative to pairwise methods that directly access reference-aligned target-view observations.

\subsubsection{Qualitative Comparison at Test Views}

\begin{figure*}[!t]
\centering
\includegraphics[width=\textwidth]{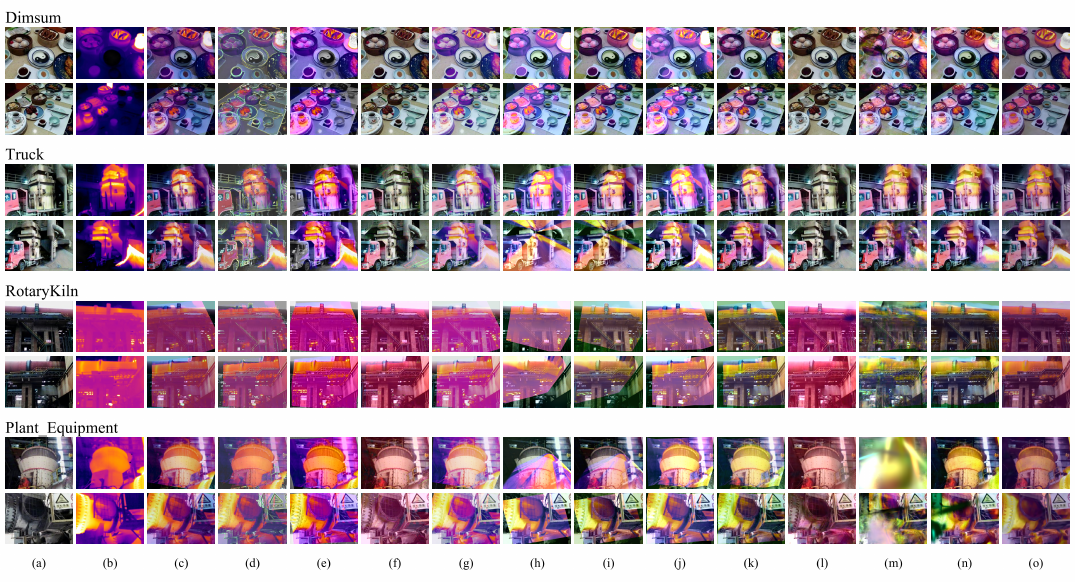}
\caption{Qualitative comparison of fusion at held-out test views, with two views from each of four RGBT-Scenes scenes. From left to right: (a) RGB reference, (b) aligned IR reference, (c) IMF, (d) MulFS-CAP, (e) C2RF, (f) HMMF, (g) FusionRegister with MMDRFuse, (h) XoFTR+CDMFusion, (i) XoFTR+Masked-DiFuser, (j) CRFT+CDMFusion, (k) CRFT+Masked-DiFuser, (l) HMMF$\rightarrow$3DGS, (m) CRFT+Masked-DiFuser$\rightarrow$3DGS, (n) CRFT+OMMG$\rightarrow$Masked-DiFuser, and (o) Ours. Pairwise methods fuse the observed RGB and warped IR images at each test view, whereas scene-level methods render from representations optimized using training views. For visual clarity, fusion results from all methods are presented using pseudocolor IR in this qualitative comparison. }
\label{fig:3}
\end{figure*}

Figure~\ref{fig:3} compares two held-out views from each of four scenes under the test protocol described above.
Pairwise methods receive RGB and residually misaligned IR observations at the test viewpoint, whereas scene-level methods render from the query viewpoint without using the test images during optimization or inference.
The RGB and aligned IR references are shown in columns (a) and (b), and pseudocolor IR is used for visualization.

The comparison highlights whether IR responses remain localized on the corresponding visible structures.
In the second Truck view, XoFTR+CDMFusion and XoFTR+Masked-DiFuser produce broad triangular color regions that cross the industrial vessel and background, unlike the localized bands on the vessel and the bright region beside the pipe in the IR reference.
Beyond3D retains these two spatially distinct responses while keeping the truck cabin and adjacent pipe recognizable.
HMMF followed by 3DGS retains a largely RGB-like appearance in this example, with little visible emphasis on the corresponding IR regions.

The scene cascades also exhibit local rendering artifacts that obscure object structure.
In the first Dimsum view, CRFT+Masked-DiFuser followed by 3DGS overlays displaced strokes on the central patterned plate and the upper-left steamer.
The same cascade renders the vessel in the first Plant\_Equipment view as diffuse color patches, while CRFT+OMMG followed by Masked-DiFuser introduces a dark green band across the circular opening in the second view.
Beyond3D keeps the plate motif and vessel outline identifiable without these large overlapping artifacts.
These examples are consistent with the motivation for coupling registration and fusion through a common scene representation, with fused attributes rendered on canonical parent geometry.

\subsubsection{Fusion Along Novel Camera Trajectories}
\begin{figure}[t]
\centering
\includegraphics[width=1.0\linewidth]{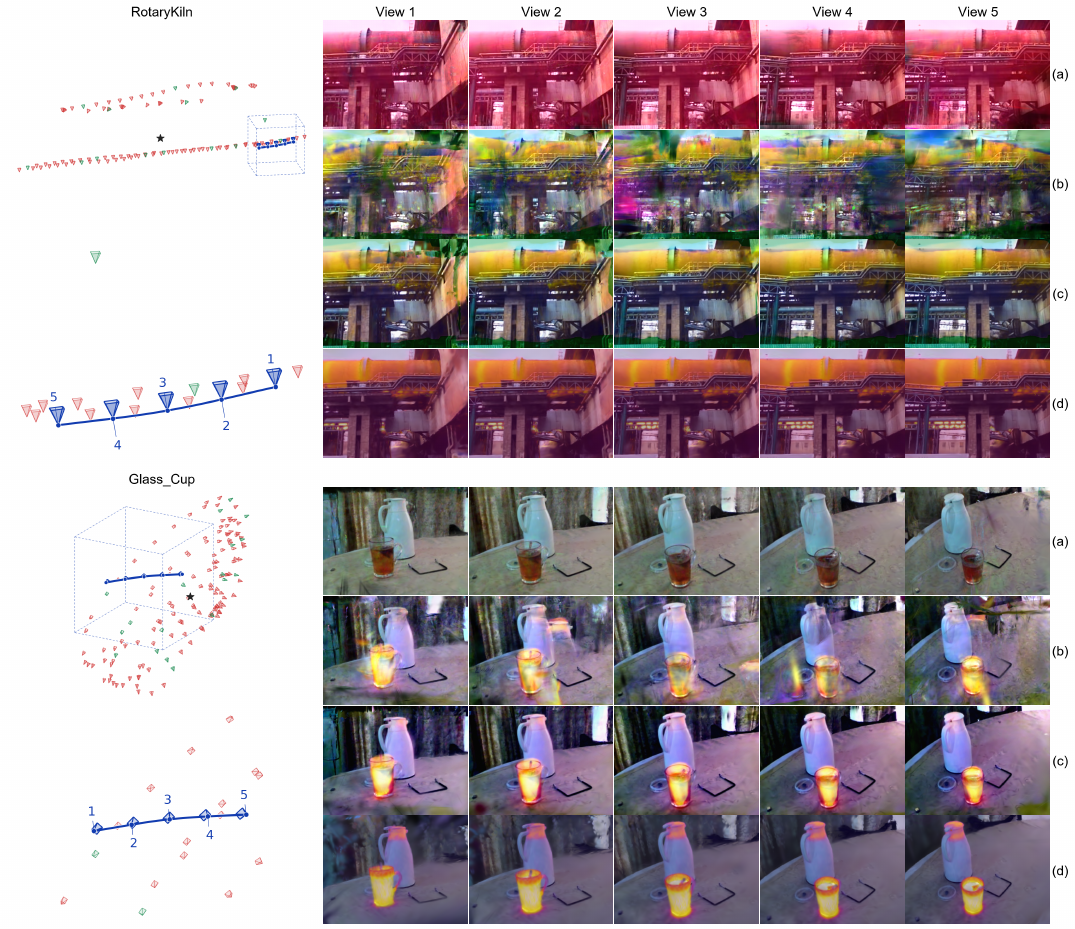}
\caption{Qualitative comparison of free-view IVF along
selected novel trajectories in RotaryKiln (top) and Glass\_Cup (bottom). The left panels show the global camera distributions and enlarged views of the selected trajectories. Red and green frusta denote training and held-out test cameras, respectively; blue frusta and curves indicate the novel query viewpoints and their trajectories. Black stars mark the COLMAP~\cite{schonberger2016structure} origin, and dashed blue boxes indicate the enlarged regions. Rows (a)--(d) correspond to FTR: HMMF$\rightarrow$3DGS, FTR: CRFT+Masked-DiFuser$\rightarrow$3DGS, RTF: CRFT+OMMG$\rightarrow$Masked-DiFuser, and Ours, respectively.}
\label{fig:4}
\end{figure}

We further examine whether spatial structure and thermal saliency remain coherent across query viewpoints along selected novel camera trajectories.
Figure~\ref{fig:4} compares five query views per scene along novel trajectories in RotaryKiln and Glass\_Cup, with identical query viewpoints and camera intrinsics across methods. In RotaryKiln, Beyond3D preserves the protruding pipe contour and localized thermal response, whereas rows (b) and (c) show duplicated or displaced structures. In Glass\_Cup, HMMF followed by 3DGS provides limited thermal emphasis, while CRFT+Masked-DiFuser followed by 3DGS produces ghosted responses around the cup and kettle. Beyond3D retains recognizable boundaries and localized thermal cues across the displayed views. These examples support spatial coherence along the selected trajectories.
View~1 of Glass\_Cup illustrates reduced fidelity far from the training viewpoints: background sharpness and fine detail deteriorate, although the cup and kettle remain recognizable with localized thermal responses.

\begin{table*}[t]
\centering
\caption{
Quantitative comparison of residual RGB--IR registration on training image pairs.
AEPE, median EPE, and P95 EPE are computed over the full-resolution valid overlap domain.
All entries are reported as scene-level mean $\pm$ standard deviation in pixels.
The best and second-best mean results within each dataset are highlighted in light red and light yellow.
}
\label{tab:residual_registration}
\setlength{\tabcolsep}{3.2pt}
\renewcommand{\arraystretch}{1.06}
\footnotesize
\begin{tabular}{l*{6}{c}}
\toprule
\multirow{2}{*}{Method}
& \multicolumn{3}{c}{RGBT-Scenes}
& \multicolumn{3}{c}{ThermoScenes}
\\
\cmidrule(lr){2-4}
\cmidrule(lr){5-7}
&
AEPE$\downarrow$
& Median EPE$\downarrow$
& P95 EPE$\downarrow$
& AEPE$\downarrow$
& Median EPE$\downarrow$
& P95 EPE$\downarrow$
\\
\midrule

\multicolumn{7}{l}{
\textit{Joint registration and fusion methods}
}\\

IMF
& $36.784 \pm 1.295$
& $34.931 \pm 1.445$
& $67.077 \pm 1.772$
& $34.207 \pm 0.751$
& $32.354 \pm 0.563$
& $62.630 \pm 1.872$
\\

MulFS-CAP
& $36.703 \pm 1.289$
& $35.209 \pm 1.445$
& $67.614 \pm 1.839$
& $35.068 \pm 0.472$
& $32.644 \pm 0.609$
& $63.506 \pm 1.877$
\\

C2RF
& \scenesecond{$29.164 \pm 2.357$}
& \scenesecond{$22.556 \pm 3.185$}
& \scenesecond{$54.330 \pm 3.963$}
& \scenesecond{$25.489 \pm 1.148$}
& \scenesecond{$17.919 \pm 0.977$}
& \scenesecond{$47.818 \pm 1.960$}
\\

HMMF
& $43.985 \pm 9.958$
& $33.743 \pm 2.623$
& $98.519 \pm 80.498$
& $42.383 \pm 12.032$
& $34.672 \pm 6.636$
& $88.276 \pm 43.592$
\\

FusionRegister
& $36.602 \pm 1.276$
& $35.056 \pm 1.452$
& $67.224 \pm 1.806$
& $34.232 \pm 0.664$
& $32.505 \pm 0.572$
& $62.972 \pm 1.920$
\\

\midrule
\multicolumn{7}{l}{
\textit{Dedicated cross-modal registration methods}
}\\

XoFTR
& $211.433 \pm 114.691$
& $187.200 \pm 120.286$
& $454.259 \pm 200.277$
& $277.470 \pm 208.877$
& $292.128 \pm 242.114$
& $457.973 \pm 259.769$
\\

CRFT
& $51.255 \pm 4.597$
& $32.445 \pm 1.870$
& $119.489 \pm 41.384$
& $46.636 \pm 0.816$
& $30.320 \pm 1.125$
& $92.323 \pm 11.344$
\\

\midrule
\textbf{Beyond3D (Ours)}
& \scenefirst{$8.750 \pm 3.743$}
& \scenefirst{$4.596 \pm 1.684$}
& \scenefirst{$9.143 \pm 5.996$}
& \scenefirst{$9.727 \pm 5.660$}
& \scenefirst{$5.725 \pm 3.238$}
& \scenefirst{$22.767 \pm 10.630$}
\\

\bottomrule
\end{tabular}%
\end{table*}

\begin{figure}[t]
\centering
\includegraphics[width=1.0\linewidth]{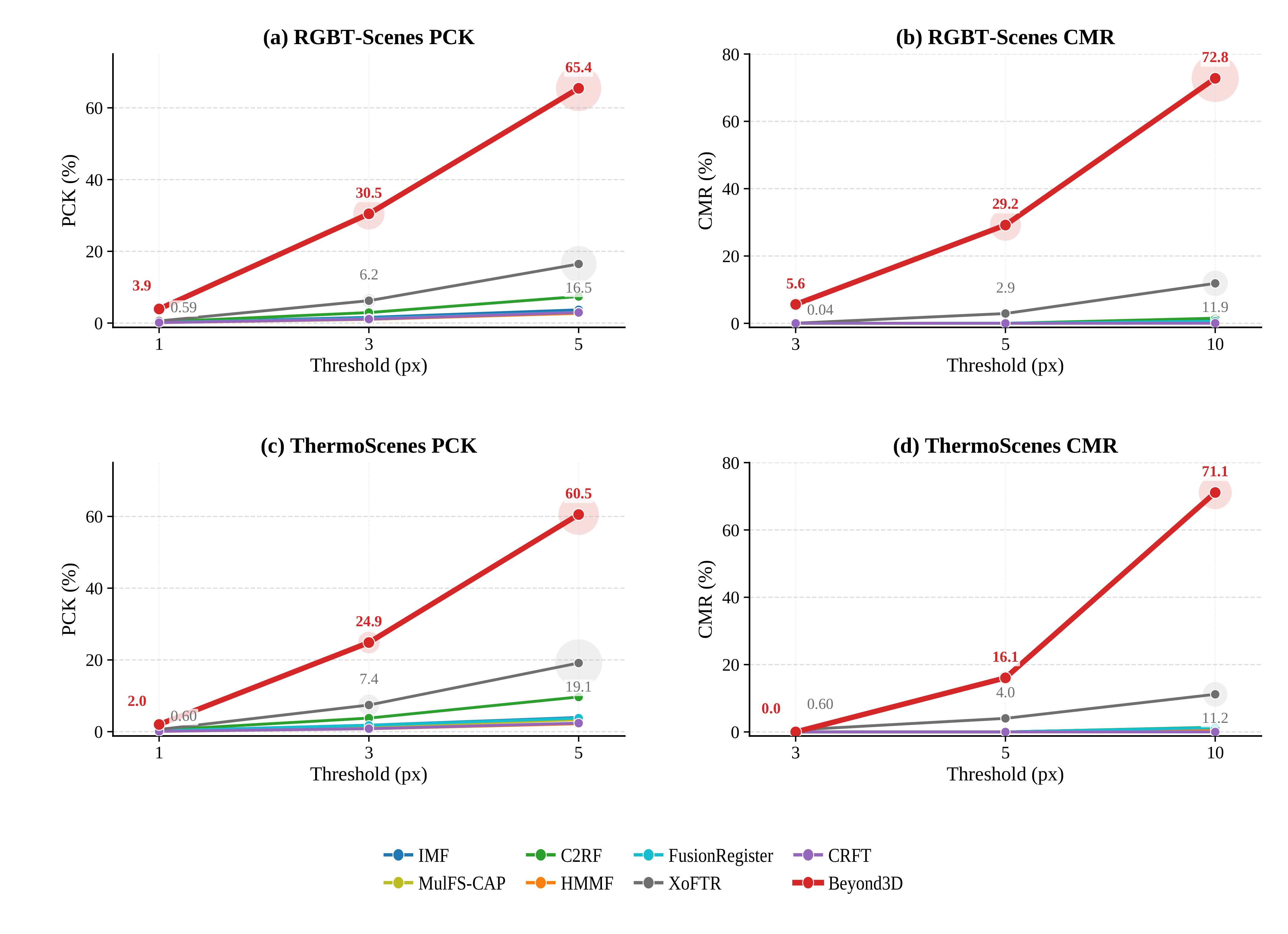}
\caption{
Full-resolution registration success curves on training pairs.
Each point denotes the scene-level mean, and the radius of the centered translucent circle is linearly scaled according to the corresponding standard deviation.
Numerical values are annotated for the two leading curves ranked at the largest threshold in each panel.
}
\label{fig:residual_registration_curves}
\end{figure}

\begin{table*}[t]
\centering
\caption{
Quantitative comparison of RGB and IR novel-view rendering at held-out viewpoints.
All entries are reported as mean $\pm$ standard deviation.
The best and second-best results are determined only among methods optimized from misaligned inputs and are highlighted in light red and light yellow.
OMMG and MMOne with reference-aligned inputs are reported separately and are not included in the ranking.
}
\label{tab:dual_modal_rendering}
\setlength{\tabcolsep}{3.2pt}
\renewcommand{\arraystretch}{1.06}
\footnotesize%
\begin{tabular}{l*{6}{c}}
\toprule
\multicolumn{7}{c}{\textbf{RGBT-Scenes}}\\
\midrule
\multirow{2}{*}{Method}
& \multicolumn{3}{c}{RGB}
& \multicolumn{3}{c}{IR}
\\
\cmidrule(lr){2-4}
\cmidrule(lr){5-7}
&
PSNR$\uparrow$
& SSIM$\uparrow$
& LPIPS$\downarrow$
& PSNR$\uparrow$
& SSIM$\uparrow$
& LPIPS$\downarrow$
\\
\midrule

\multicolumn{7}{l}{
\textit{Methods optimized from residually misaligned inputs}
}\\

HMMF$\rightarrow$3DGS
& $15.913 \pm 5.052$
& $0.700 \pm 0.114$
& $0.405 \pm 0.153$
& $9.157 \pm 1.520$
& $0.354 \pm 0.130$
& $0.773 \pm 0.081$
\\

CRFT+Masked-DiFuser$\rightarrow$3DGS
& $13.531 \pm 1.376$
& $0.587 \pm 0.081$
& $0.506 \pm 0.076$
& $8.628 \pm 0.888$
& $0.317 \pm 0.083$
& $0.785 \pm 0.064$
\\

CRFT+OMMG
& $22.847 \pm 2.366$
& $0.809 \pm 0.056$
& \scenesecond{$0.231 \pm 0.041$}
& \scenesecond{$15.088 \pm 2.527$}
& \scenesecond{$0.668 \pm 0.082$}
& \scenesecond{$0.391 \pm 0.094$}
\\

CRFT+MMOne
& \scenefirst{$24.202 \pm 2.645$}
& \scenefirst{$0.833 \pm 0.060$}
& \scenefirst{$0.218 \pm 0.053$}
& $14.482 \pm 2.618$
& $0.660 \pm 0.081$
& $0.425 \pm 0.106$
\\

\textbf{Beyond3D (Ours)}
& \scenesecond{$24.145 \pm 2.996$}
& \scenesecond{$0.832 \pm 0.074$}
& $0.260 \pm 0.076$
& \scenefirst{$23.451 \pm 2.566$}
& \scenefirst{$0.829 \pm 0.086$}
& \scenefirst{$0.147 \pm 0.077$}
\\

\midrule
\multicolumn{7}{l}{
\textit{Aligned-input references}
}\\

OMMG (aligned)
& $24.330 \pm 2.458$
& $0.839 \pm 0.057$
& $0.190 \pm 0.047$
& $24.679 \pm 3.119$
& $0.864 \pm 0.056$
& $0.117 \pm 0.047$
\\

MMOne (aligned)
& $24.797 \pm 2.599$
& $0.847 \pm 0.057$
& $0.199 \pm 0.051$
& $24.622 \pm 3.172$
& $0.867 \pm 0.056$
& $0.122 \pm 0.048$
\\

\specialrule{0.8pt}{3pt}{2pt}
\multicolumn{7}{c}{\textbf{ThermoScenes}}\\
\midrule
\multirow{2}{*}{Method}
& \multicolumn{3}{c}{RGB}
& \multicolumn{3}{c}{IR}
\\
\cmidrule(lr){2-4}
\cmidrule(lr){5-7}
&
PSNR$\uparrow$
& SSIM$\uparrow$
& LPIPS$\downarrow$
& PSNR$\uparrow$
& SSIM$\uparrow$
& LPIPS$\downarrow$
\\
\midrule

\multicolumn{7}{l}{
\textit{Methods optimized from misaligned inputs}
}\\

HMMF$\rightarrow$3DGS
& $10.437 \pm 3.678$
& $0.495 \pm 0.210$
& $0.626 \pm 0.193$
& $7.290 \pm 0.881$
& $0.528 \pm 0.067$
& $0.928 \pm 0.022$
\\

CRFT+Masked-DiFuser$\rightarrow$3DGS
& $12.174 \pm 3.002$
& $0.452 \pm 0.232$
& $0.573 \pm 0.165$
& $7.505 \pm 1.747$
& $0.462 \pm 0.101$
& $0.898 \pm 0.017$
\\

CRFT+OMMG
& \scenesecond{$25.504 \pm 7.594$}
& $0.793 \pm 0.066$
& \scenesecond{$0.241 \pm 0.064$}
& \scenesecond{$17.676 \pm 1.153$}
& $0.862 \pm 0.013$
& $0.342 \pm 0.048$
\\

CRFT+MMOne
& \scenefirst{$27.625 \pm 6.688$}
& \scenefirst{$0.848 \pm 0.062$}
& \scenefirst{$0.179 \pm 0.058$}
& $17.557 \pm 0.677$
& \scenesecond{$0.893 \pm 0.006$}
& \scenesecond{$0.318 \pm 0.058$}
\\

\textbf{Beyond3D (Ours)}
& $25.485 \pm 5.671$
& \scenesecond{$0.818 \pm 0.062$}
& $0.243 \pm 0.069$
& \scenefirst{$31.758 \pm 2.143$}
& \scenefirst{$0.963 \pm 0.018$}
& \scenefirst{$0.122 \pm 0.085$}
\\

\midrule
\multicolumn{7}{l}{
\textit{Aligned-input references}
}\\

OMMG (aligned)
& $27.296 \pm 6.930$
& $0.832 \pm 0.068$
& $0.189 \pm 0.061$
& $31.680 \pm 2.097$
& $0.972 \pm 0.006$
& $0.098 \pm 0.033$
\\

MMOne (aligned)
& $28.661 \pm 6.643$
& $0.850 \pm 0.061$
& $0.178 \pm 0.059$
& $32.396 \pm 1.995$
& $0.977 \pm 0.003$
& $0.086 \pm 0.024$
\\

\bottomrule
\end{tabular}%

\end{table*}

\subsection{Analysis of Supporting Capabilities}

\subsubsection{Residual Registration} 
We evaluate residual cross-modal registration on the training RGB--IR pairs, where the known perturbations provide dense ground-truth correspondences.
The comparison includes IMF, MulFS-CAP, C2RF, HMMF, and FusionRegister as joint registration and fusion methods, XoFTR as matching-based registrar, and CRFT as dense-flow registrar.
The correspondences predicted by the matching-based method are converted into homographies using RANSAC.
For a unified comparison, all metrics are computed over the full-resolution valid overlap.
Table~\ref{tab:residual_registration} reports AEPE, median EPE, and P95 EPE, while Fig.~\ref{fig:residual_registration_curves} reports PCK and CMR under increasing pixel thresholds.
PCK measures the percentage of valid pixels whose endpoint error does not exceed the threshold, while CMR measures the percentage of image pairs whose AEPE satisfies the threshold.

Because Beyond3D jointly optimizes a view-specific transformation with the scene representation for each observed training view, whereas the compared methods process each image pair independently without scene-level refinement, this experiment evaluates the final residual alignment supplied to scene optimization rather than registration generalization to unseen pairs.
The comparison should therefore be interpreted under the native operating regime of each method rather than as a compute-matched benchmark of standalone registrars.

As shown in Table~\ref{tab:residual_registration}, Beyond3D achieves the lowest AEPE, median EPE, and P95 EPE on both datasets.
Relative to C2RF, the strongest competing method in terms of aggregate errors, it reduces the three metrics by $70.0\%$, $79.6\%$, and $83.2\%$ on RGBT-Scenes, and by $61.8\%$, $68.1\%$, and $52.4\%$ on ThermoScenes.
The success curves in Fig.~\ref{fig:residual_registration_curves} further show that these gains extend across different error tolerances.
On RGBT-Scenes, Beyond3D reaches $65.4\%$ PCK at $5$ pixels and $72.8\%$ CMR at $10$ pixels, compared with $16.5\%$ and $11.9\%$ for the next-best XoFTR.
On ThermoScenes, it achieves $60.5\%$ PCK at $5$ pixels and $71.1\%$ CMR at $10$ pixels, substantially exceeding the corresponding XoFTR results of $19.1\%$ and $11.2\%$.
XoFTR retains a marginal advantage under the strict $3$-pixel CMR threshold on ThermoScenes, while Beyond3D becomes markedly stronger at the more permissive thresholds.
These results do not imply superiority as a feed-forward pairwise registrar.
Rather, they demonstrate that jointly exploiting multi-view scene evidence through iterative per-view refinement yields substantially more accurate optimized residual registration on the training observations.

\subsubsection{Dual-Modal Rendering Quality}
We examine whether the learned representation improves IR rendering while retaining RGB fidelity, as motivated by the separation of canonical structure and modality-specific adaptation.
The RGB and IR branches are rendered independently at held-out viewpoints and evaluated using PSNR, SSIM, and LPIPS~\cite{zhang2018unreasonable} against test images excluded from optimization.
OMMG and MMOne~\cite{gu2025mmone} trained on reference-aligned observations serve as contextual references and are excluded from ranking.

As reported in Table~\ref{tab:dual_modal_rendering}, Beyond3D attains the best IR PSNR, SSIM, and LPIPS among the methods trained on misaligned observations on both datasets.
Relative to the strongest competing result for each metric, IR PSNR increases by 8.36~dB on RGBT-Scenes and 14.08~dB on ThermoScenes, while IR LPIPS decreases by $62.4\%$ and $61.6\%$, respectively.
RGB reconstruction remains competitive, although its preservation depends on the metric and dataset.
On RGBT-Scenes, RGB PSNR and SSIM are only 0.057~dB and 0.001 below CRFT+MMOne, respectively, whereas LPIPS is higher, at 0.260 versus 0.218.
On ThermoScenes, RGB PSNR and LPIPS remain close to CRFT+OMMG, with higher SSIM, while CRFT+MMOne remains stronger across three RGB metrics.

\begin{figure}[!t]
\centering
\includegraphics[width=\columnwidth]{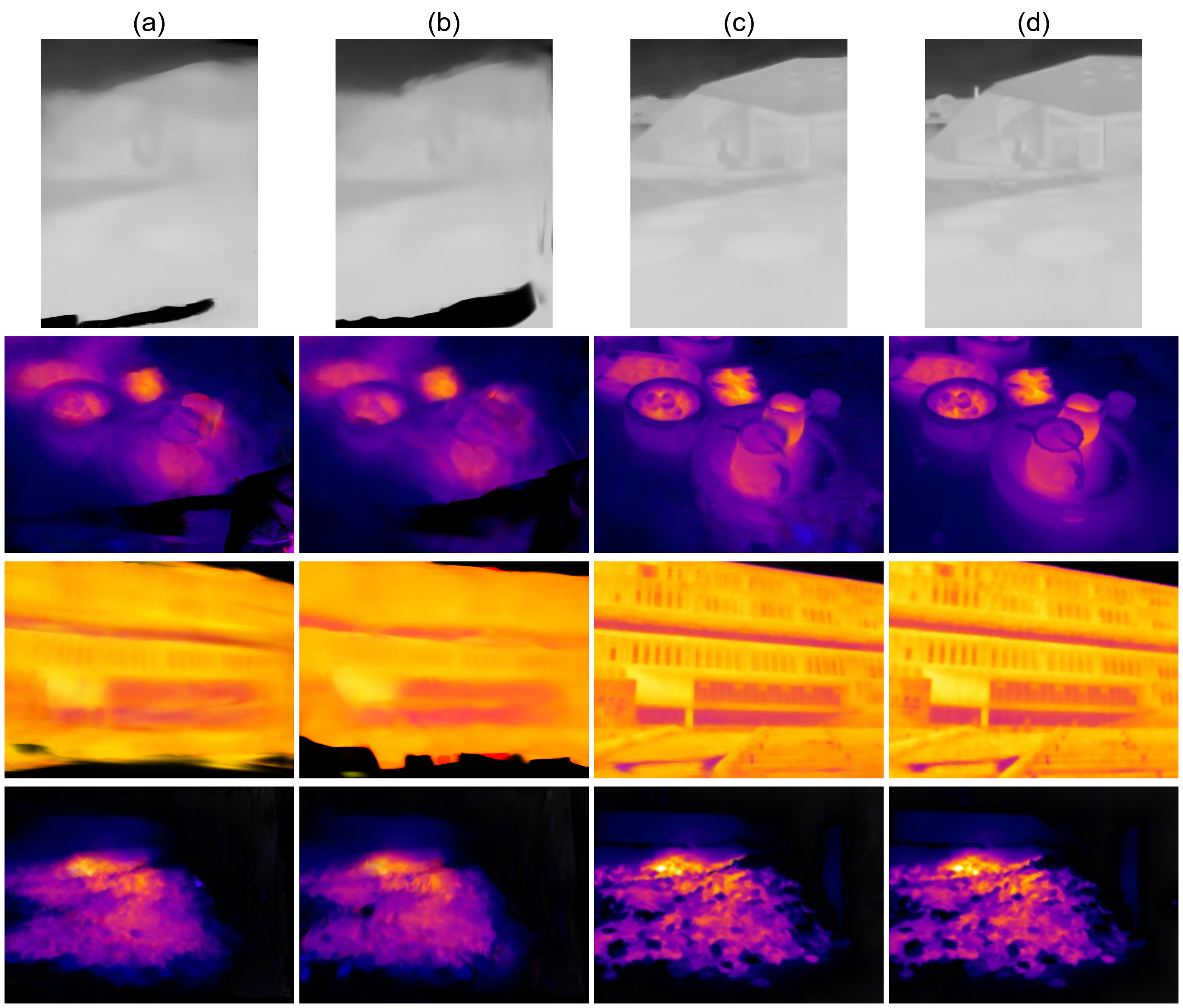}
\caption{Qualitative comparison of novel-view IR rendering from residually misaligned RGB-IR training observations. Columns (a), (b), (c), and (d) show CRFT+OMMG, CRFT+MMOne, Ours, and the ground-truth IR reference, respectively. Each row compares the same held-out viewpoint.}
\label{fig:fig_nvs}
\end{figure}

Figure~\ref{fig:fig_nvs} illustrates the improvement in IR reconstruction through identifiable local structures.
CRFT+OMMG and CRFT+MMOne blur the roof contour in the first row and the repeated facade structures in the third row, while Beyond3D reproduces these features more distinctly and reduces the large dark artifacts along the lower image boundaries.
In the second row, the cup rims and kettle contour are better separated from the surrounding tabletop, and the bottom row shows more distinguishable boundaries between individual foreground elements instead of a diffuse thermal region.
These observations are consistent with the intended role of modality-specific children in representing local IR structure around a common scene anchor.
Residual differences from the reference remain, particularly around the tabletop objects.

\subsection{Ablation Study}

\subsubsection{Core Component Ablation}
\label{sec:core_component_ablation}
We evaluate cumulative stages and five component variants against the full model across all 18 scenes with equal scene weights (Table~\ref{tab:core_component_ablation}).
We report AEPE on training pairs, RGB and IR PSNR for rendering at all held-out test views, and SCD and $\mathrm{SSIM}_d$ for fusion at the same views.

We remove the FRC reference branch or replace all alternating updates with simultaneous updates throughout Stages~2 and~3.
Both variants retrain the coarse representation, with the no-FRC variant retaining Full's Stage~1 initialization.

The other three variants share Full's Stage~2 checkpoint.
They fix child displacement, scale, and opacity residuals at zero, jointly remove Gaussian–ray correspondence, or disable RGB-protected gradient correction.
The geometry variant retains both modality branches and trainable parent geometry but reduces geometric degrees of freedom.
The gradient variant retains both reconstruction objectives.

\begin{table}[t]
\centering
\caption{Core component ablations aggregated over all scenes with equal weights.
Full denotes Stages~1+2+3.
$\dagger$ marks half-resolution PSNR for Stages~1+2.}
\label{tab:core_component_ablation}
\footnotesize
\setlength{\tabcolsep}{5pt}
\renewcommand{\arraystretch}{1.12}
\begin{adjustbox}{max width=\columnwidth}
\begin{tabular}{l*{5}{c}}
\toprule
Configuration
& \makecell{AEPE\\(px)$\downarrow$}
& \makecell{RGB PSNR\\(dB)$\uparrow$}
& \makecell{IR PSNR\\(dB)$\uparrow$}
& SCD$\uparrow$ & $\mathrm{SSIM}_d\uparrow$ \\
\midrule
Stage 1
& 87.929 & / & / & / & / \\
Stages 1+2$^{\dagger}$
& 20.954 & 22.027 & 20.817 & / & / \\
\textbf{Full (Stages 1+2+3)}
& 8.967 & 24.443 & 25.297 & 1.440 & 0.612 \\
\midrule
w/o FRC
& 32.015 & 17.500 & 16.138 & 1.348 & 0.493 \\
w/o alternating optimization
& 56.696 & 15.852 & 15.230 & 1.255 & 0.451 \\
w/o parent-child geometry
& 12.224 & 23.537 & 23.708 & 1.412 & 0.573 \\
w/o Gaussian–ray correspondence
& 9.772 & 24.705 & 25.210 & 1.119 & 0.506 \\
w/o RGB-protected gradient
& 10.319 & 23.451 & 23.488 & 1.371 & 0.474 \\
\bottomrule
\end{tabular}
\end{adjustbox}
\end{table}

Table~\ref{tab:core_component_ablation} supports progressive registration refinement and highlights the importance of controlling registration and representation co-adaptation.
The concurrent degradation in alignment and dual-modal reconstruction when FRC or alternation is removed from both refinement stages.
Fixing the child geometric residuals at zero weakens reconstruction and fusion, supporting bounded spatial adaptation to modality differences.
Removing RGB protection also degrades both modalities and fusion, suggesting that controlling conflicting updates to shared parent parameters can benefit their common representation.
The correspondence ablation exposes a distinction between reconstruction fidelity and fusion quality.
Removing Gaussian–ray correspondence slightly improves RGB reconstruction and changes IR reconstruction little, yet degrades fusion.
This pattern supports their combined role in primitive-level fusion, where paired attributes must remain compatible when combined on canonical parent geometry.

\subsubsection{Effect of Parent-Child Representation Topology}
\label{sec:topology_ablation}
We compare the canonical-parent model with six variants in which RGB or IR is represented directly by the parent and the other modality by $K\in\{1,2,4\}$ children.
All six variants start Stage~3 from the same Stage~2 output, with identical scene representations and registration parameters.

All variants use pre-projection fusion.
Opposite-modality child colors are aggregated using their Gaussian responses at the parent center.
The resulting local appearance is blended with the parent-modality appearance using the view-conditioned IR weight in Eq.~\eqref{eq:fusion_weight}, then rendered on the parent geometry.
At fixed $K$, the two parent choices have matched primitive counts, while increasing $K$ also increases capacity.
Table~\ref{tab:topology_ablation} reports the final Stage~3 registration, reconstruction, and fusion metrics.

\begin{table}[t]
\centering
\caption{Ablation study on Parent-Child Representation Topology.
$K_{\mathrm{ir}}$ and $K_{\mathrm{rgb}}$ denote IR and RGB children per parent.}
\label{tab:topology_ablation}
\footnotesize
\setlength{\tabcolsep}{2.5pt}
\renewcommand{\arraystretch}{1.12}
\begin{adjustbox}{max width=\columnwidth}
\begin{tabular}{@{}l*{5}{c}@{}}
\toprule
Configuration
& \makecell{AEPE\\(px)$\downarrow$}
& \makecell{RGB PSNR\\(dB)$\uparrow$}
& \makecell{IR PSNR\\(dB)$\uparrow$}
& SCD$\uparrow$ & $\mathrm{SSIM}_d\uparrow$ \\
\midrule
\textbf{Full} & 8.967 & 24.443 & 25.297 & 1.440 & 0.612 \\
\midrule
\multicolumn{6}{l}{\textit{RGB parent}} \\
$K_{\mathrm{ir}}=1$ & 9.734 & 24.356 & 24.736 & 1.379 & 0.599 \\
$K_{\mathrm{ir}}=2$ & 10.915 & 24.342 & 24.809 & 1.353 & 0.582 \\
$K_{\mathrm{ir}}=4$ & 12.490 & 24.081 & 25.855 & 1.284 & 0.567 \\
\midrule
\multicolumn{6}{l}{\textit{IR parent}} \\
$K_{\mathrm{rgb}}=1$ & 10.316 & 24.187 & 24.786 & 1.365 & 0.572 \\
$K_{\mathrm{rgb}}=2$ & 11.510 & 24.003 & 24.318 & 1.326 & 0.589 \\
$K_{\mathrm{rgb}}=4$ & 12.927 & 23.643 & 23.579 & 1.232 & 0.561 \\
\bottomrule
\end{tabular}
\end{adjustbox}
\end{table}

In Table~\ref{tab:topology_ablation}, Full leads on all reported metrics except IR PSNR, which is highest for the RGB-parent variant with $K_{\mathrm{ir}}=4$.
For the RGB-parent variants, increasing the number of IR children improves IR reconstruction but worsens registration and fusion.
This divergence shows that additional capacity for one modality does not necessarily improve alignment or fusion on the parent geometry.
For the IR-parent variants, increasing the number of RGB children degrades registration, both reconstructions, and SCD, while $\mathrm{SSIM}_d$ peaks at $K_{\mathrm{rgb}}=2$ and remains below Full.
These patterns favor the proposed allocation of shared and modality-specific geometry.

\subsubsection{Effect of Modality-Specific Geometric Degrees of Freedom}
\label{sec:geometric_dof_ablation}
We examine whether the proposed geometric parameterization balances reconstruction and fusion quality with storage cost.
Starting from the same Stage~2 initialization, we retain one RGB child and one IR child per canonical parent and compare the proposed configuration with four variants, each changing one geometric factor.
$P$ enables an additional bounded displacement of the pair center, while $R$ enables independent local rotation residuals for the two modalities.
$S$ and $O$ select modality-specific scale and opacity residuals, respectively, with zero indicating a shared residual.
The proposed $(P,R,S,O)=(0,0,1,1)$ configuration therefore retains symmetric child centers and the parent rotation while allowing modality-specific scales and opacities.

The $1011$ and $0111$ variants test whether relaxing the center and rotation constraints yields sufficient quality gains to justify increased storage cost.
The $0010$ and $0001$ variants test the effects of sharing opacity and scale residuals, respectively.
We report RGB and IR PSNR together with SCD and $\mathrm{SSIM}_d$ at held-out views.
The archived models contain approximately $213{,}963$ Gaussians per scene on average, comprising $71{,}321$ canonical parents and $142{,}642$ modality-specific children.
Storage cost is defined as the complete Stage~3 checkpoint size.
Table~\ref{tab:geometric_dof_ablation} reports storage cost together with reconstruction and fusion quality.

\begin{table}[t]
\centering
\caption{Geometric degrees of freedom.
Storage costs are averaged equally over all scenes and reported in decimal MB.}
\label{tab:geometric_dof_ablation}
\footnotesize
\setlength{\tabcolsep}{2.5pt}
\renewcommand{\arraystretch}{1.12}
\begin{adjustbox}{max width=\columnwidth}
\begin{tabular}{@{}c*{5}{c}@{}}
\toprule
$PRSO$
& \makecell{Storage cost\\(MB)$\downarrow$}
& \makecell{RGB PSNR\\(dB)$\uparrow$}
& \makecell{IR PSNR\\(dB)$\uparrow$}
& SCD$\uparrow$ & $\mathrm{SSIM}_d\uparrow$ \\
\midrule
\textbf{0011} & 79.136 & 24.443 & 25.297 & 1.440 & 0.612 \\
1011 & 80.847 & 24.504 & 25.446 & 1.403 & 0.661 \\
0111 & 82.559 & 24.276 & 24.945 & 1.433 & 0.599 \\
0010 & 78.565 & 23.925 & 24.104 & 1.388 & 0.574 \\
0001 & 77.424 & 23.757 & 24.241 & 1.339 & 0.560 \\
\bottomrule
\end{tabular}
\end{adjustbox}
\end{table}

Table~\ref{tab:geometric_dof_ablation} supports retaining modality-specific scale and opacity residuals.
Sharing either yields modest storage savings but weakens both reconstruction and fusion, consistent with the role of these residuals in accommodating modality-dependent spatial support.
Independent rotation residuals increase storage without improving any reported quality metric.
Relaxing the center constraint improves both rendering PSNRs and $\mathrm{SSIM}_d$, but lowers SCD and increases storage.
The proposed $0011$ retains the highest SCD with less storage than the expanded variants, supporting its use as a compromise.

\subsubsection{Stability Across Random Runs}
\label{sec:random_run_stability}
\begin{figure}[t]
\centering
\includegraphics[width=\linewidth]{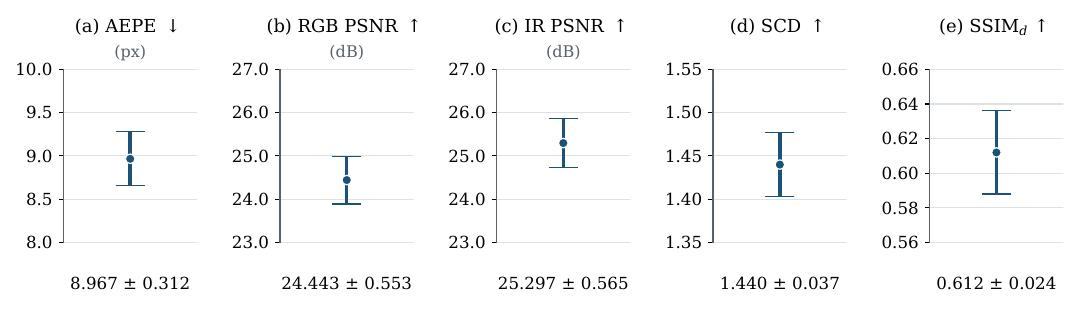}
\caption{Variability across ten independent runs of the complete pipeline.
Markers indicate means of scene-averaged run scores, and error bars span a standard deviation in each direction.}
\label{fig:random_run_stability}
\end{figure}

We assess sensitivity to randomness through ten independent runs of the complete pipeline, repeating Stages~1, 2, and~3 in each run.
We evaluate registration using AEPE on training pairs, and reconstruction and fusion using RGB and IR PSNR, SCD, and $\mathrm{SSIM}_d$ at held-out views.
For each metric, we average scene-level scores with equal weights within each run, then compute the mean and sample standard deviation across the ten run-level scores.
Figure~\ref{fig:random_run_stability} reports these statistics as mean $\pm$ one sample standard deviation.
The observed dispersion suggests repeatable aggregate performance under the evaluated protocol.

\section{Conclusion}
\label{sec:conclusion}

\subsection{Conclusion}
This work extends pairwise IVF to scene-level fused rendering from arbitrary user-defined viewpoints. We developed a coarse-to-fine framework that jointly resolves residual registration, multimodal scene representation, and view-dependent fusion. By stabilizing a canonical scene before accommodating bounded modality-specific spatial variation, the proposed method enables coherent fused rendering at both observed and arbitrary user-defined viewpoints while maintaining competitive fusion quality at captured views.
More broadly, free-view IVF provides a flexible interface for exploring complementary visible and thermal information in three-dimensional space. It can support applications in which users or autonomous systems need to revisit thermal anomalies, inspect their spatial relationship with visible structures, or continuously observe a scene along virtual trajectories, such as aerial inspection, robotic scanning, and intelligent monitoring. We hope this perspective encourages further research toward dynamic, generalizable, and large-scale multimodal scene modeling, and helps connect infrared--visible fusion with the broader development of multimodal 3D vision.

\subsection{Limitations and Future Work}
Despite its effectiveness, the present method has several limitations. 
First, it assumes a static scene and therefore cannot account for moving objects or temporal changes in observations. Extending free-view IVF to dynamic scenes is substantially more challenging, as cross-modal registration, scene representation, and fusion must be jointly resolved across both space and time. 
Second, our method relies on per-scene optimization, rather than providing direct cross-scene generalization. This limitation is largely inherited from optimization-based neural scene representations. Emerging feed-forward 3D Gaussian models~\cite{charatan2024pixelsplat,chen2024mvsplat} provide a promising route toward efficient and generalizable free-view IVF, but these methods demand large\nobreakdash-scale, high\nobreakdash-quality multi-view RGB--IR 3D datasets. Constructing a larger and more diverse RGB--IR 3D benchmark would enable more comprehensive evaluation and provide a valuable resource for the broader multimodal 3D vision community.


\end{document}